\documentclass[acmsmall]{acmart}
\usepackage[normalem]{ulem}

\AtBeginDocument{%
  \providecommand\BibTeX{{%
    \normalfont B\kern-0.5em{\scshape i\kern-0.25em b}\kern-0.8em\TeX}}}

\setcopyright{acmcopyright}
\copyrightyear{2024}
\acmYear{2024}
\acmDOI{XXXXXXX.XXXXXXX}

\acmConference[CSCW '24]{The 27th ACM Conference on Computer-Supported Cooperative Work and Social Computing}{November 09--13,
  2024}{San José, Costa Rica}
\acmPrice{15.00}
\acmISBN{978-1-4503-XXXX-X/18/06}

\acmSubmissionID{2085}

\begin{document}
\title{Race and Privacy in Broadcast Police Communications}
\author{Pranav Narayanan Venkit}
\email{pnv5011@psu.edu}
\affiliation{%
  \institution{Pennslylvania State University}
  \city{University Park}
  \state{Pennsylvania}
  \country{USA}
  \postcode{16802}
}

\author{Christopher Graziul}
\email{graziul@uchicago.edu}
\affiliation{%
  \institution{University of Chicago}
  \city{Chicago}
  \state{Illinois}
  \country{USA}
}

\author{Miranda Ardith Goodman}
\email{mag6785@psu.edu}
\affiliation{%
  \institution{Pennslylvania State University}
  \city{University Park}
  \state{Pennsylvania}
  \country{USA}
  \postcode{16802}
}

\author{Samantha Nicole Kenny}
\email{snk5300@psu.edu}
\affiliation{%
  \institution{Pennslylvania State University}
  \city{University Park}
  \state{Pennsylvania}
  \country{USA}
  \postcode{16802}
}

\author{Shomir Wilson}
\email{shomir@psu.edu}
\affiliation{%
  \institution{Pennslylvania State University}
  \city{University Park}
  \state{Pennsylvania}
  \country{USA}
  \postcode{16802}
}






\renewcommand{\shortauthors}{Narayanan Venkit, et al.}

\begin{abstract}


Radios are essential for the operations of modern police departments, and they function as both a collaborative communication technology and a sociotechnical system. However, little prior research has examined their usage or their connections to individual privacy and the role of race in policing, two growing topics of concern in the US. As a case study, we examine the Chicago Police Department's (CPD's) use of broadcast police communications (BPC) to coordinate the activity of law enforcement officers (LEOs) in the city. From a recently assembled archive of $80,775$ hours of BPC associated with CPD operations, we analyze human-generated text transcripts of radio transmissions broadcast 9:00 AM to 5:00 PM on August 10th, 2018 in one majority Black, one majority white, and one majority Hispanic area of the city (24 hours of audio) to explore three research questions: (1) Do BPC reflect reported racial disparities in policing? (2) How and when is gender, race/ethnicity, and age mentioned in BPC? (3) To what extent do BPC include sensitive information, and who is put at most risk by this practice? (4) To what extent can large language models (LLMs) heighten this risk? We explore the vocabulary and speech acts used by police in BPC, comparing mentions of personal characteristics to local demographics, the personal information shared over BPC, and the privacy concerns that it poses. Analysis indicates (a) policing professionals in the city of Chicago exhibit disproportionate attention to Black members of the public regardless of context, (b) sociodemographic characteristics like gender, race/ethnicity, and age are primarily mentioned in BPC about event information, and (c) disproportionate attention introduces disproportionate privacy risks for Black members of the public. This study shows BPC can provide a novel window into disproportionate attention (i.e., via radio communications) by law enforcement officers to specific racial groups, leading to increased privacy vulnerability for those groups, particularly Black males. 

\end{abstract}


\received{18 July 2023}
\received[revised]{16 January 2024}
\received[accepted]{? June 2024}

\maketitle

\section{Introduction}
Modern police radio communication networks are sociotechnical systems that coordinate operations for thousands of law enforcement officers using sophisticated electronics and extensive verbal protocols. Among communication networks, they are notable for their relevance to policing the physical world. However, these networks are vulnerable to privacy breaches and can expose sensitive personal information to unintended listeners \cite{vargas2019digital}. Moreover, research shows the study of digital technologies used by law enforcement agencies can reveal previously unknown racial disparities in the policing of certain sociodemographic groups, which can exacerbate existing social inequalities linked to race/ethnicity \cite{vargas2019digital, voigt2017language, rho_escalated_2023}.\footnote{We use the term race/ethnicity in conformity with how the Census Bureau and Bureau of Justice Statistics \cite{bureau_of_justice_statistics_raceethnicity_2021} differentiates Hispanic ethnicity from race. We also use the term ``Black" to refer to ``Black or African American" for brevity.} 

We scrutinize language patterns in broadcast police communication (BPC) used by police dispatchers and law enforcement officers employed by the City of Chicago, with the objective of uncovering potential disparities in police attention to specific demographic groups. BPC reflect policing behavior \cite{wells1997research}, and recent research suggests they exhibit racial disparities in privacy vulnerability \cite{vargas2019digital}. By examining BPC, we gain insight into how policing behavior manifests through interactions with these technologies. Using transcribed text data extracted from CPD-related BPC, we inspect the language employed between dispatchers and officers in three \textit{dispatch zones}, distinct geographic regions used to coordinate police activity in the city of Chicago, over an 8 hour period (9:00 AM to 5:00 PM) on August 10th, 2018. 

Our approach involves a combination of lexical analysis and thematic qualitative coding, which enabled us to understand the general purpose of each communication and evaluate the existence of disparity regarding attention and privacy vulnerability of conversations in BPC. Our study further employs the lens of assessing the effectiveness of large language models to comprehend how privacy vulnerability is heightened if text transcripts of such communications are available. We explore the consequences of using this technology to extract sensitive personal data. We build upon the analysis showing that existing language models can infer personal information from text or audio data \cite{carlini2021extracting, staab2023beyond}, highlighting its broader implications for privacy in communicative and collaborative technologies.

Our analysis supports four related goals: (1) validate BPC exhibit previously reported racial disparities in policing using mentions of race/ethnicity as an indicator of police attention; (2) describe patterns in privacy vulnerability within BPC; (3) assess BPC-related privacy vulnerability across racial/ethnic groups; (4) test how BPC privacy risks scale given emerging technologies. These goals correspond to the following research questions: 

\begin{itemize}
    
    \item \textit{Do radio transmissions between policing professionals reflect reported racial disparities in policing?}  

    \item \textit{How and when are gender, race/ethnicity, and age mentioned in radio transmissions between policing professionals?} 
    
    \item \textit{To what extent does BPC include sensitive information and who is put at most risk by this practice?} 

    \item \textit{Can privacy vulnerability in BPC be heightened and misused by the use of pre-trained Large Language Models (LLMs)?}
    
\end{itemize}

Overall, results demonstrate a disproportionate attention by law enforcement officers toward specific racial groups, even relative to the demographic makeup of each area, indicating an increase in privacy vulnerability for those groups. This finding is important since BPC occur prior to police assessment of criminal behavior, and thus disproportionate attention precedes law enforcement activity (e.g., assessment whether a crime has possibly occurred). This order of events indicates demographic makeup of communities is more appropriate than demographic makeup of crime reports for assessing disparities: The principles of procedural justice require officers show neutrality in decision making concerning encounters with members of the public \cite{solum_procedural_2004} and adherence to these principles is directly related to public trust in police \cite{gau_procedural_2010, mazerolle_legitimacy_2013, mazerolle_shaping_2013, weisburd_reforming_2022}, but race/ethnicity (and related potential stereotyping) may be one of few inputs to officer decision making (e.g., threat assessment) and the disproportionate representation of groups in BPC would evidence a systemic lack of neutrality in deciding who to monitor and, ultimately, engage. Disproportionality in BPC may arise due to disproportionate reports of crime involving specific sociodemographic groups~\cite{gillooly_how_2024}, but this does not negate the disparity in police attention.

Our study demonstrates the importance of analyzing police radio communication to reveal potential disparities in communication between law enforcement personnel about specific sociodemographic groups, highlighting the presence of privacy vulnerabilities in BPC. Our analysis, coupled with the use of current NLP technologies, such as LLMs, shows the repercussions of vulnerabilities within BPC. This highlights the imperative to acknowledge and address these vulnerabilities inherent in such collaborative systems
and provides insights into the language used by policing professionals in two-way radio communication.

\section{Background}

Radio broadcasts have been used as a policing technology and tool for over 90 years \cite{andrews_greatest_2021, ieee_1987_5527521}. The emergence of police radio communications in the United States can be traced to systems in Detroit, Michigan (1928, one-way broadcast) and Bayonne, New Jersey (1932, two-way transmissions) \cite{ieee_1987_5527521}. Development and adoption of these systems began in the early 1920s, progressed quickly over the following two decades, and spread to communities large and small \cite{poli_development_1942}. Two-way radio systems rely on the broadcast of messages from a mobile or stationary transmitter to multiple receivers simultaneously, enabling fast dissemination of information to listeners. 

Today, two-way radio communication is vital for law enforcement operations around the world, facilitating both coordination and information sharing between policing professionals. Until the 21st century, most agencies in the United States used unencrypted radio frequencies to communicate, raising concerns that such sociotechnical systems introduce privacy risks to those being policed \cite{vargas2019digital}. Given the routine sharing of sensitive information to coordinate police activity, and substantial evidence of racial disparities in policing \cite{goncalves_few_2021, ang_effects_2021, edwards_risk_2019, aggarwal_high-frequency_2022, nix_birds_2017, pierson_large-scale_2020, franchi_detecting_2023}, it is logical to ask: What potential disparities do radio-based law enforcement communication systems introduce, perpetuate, or otherwise reveal? 

Identification of disparities in policing via BPC helps enhance police transparency as well as protect the civil liberties of those being policed, if only through identifying instances of potential civil rights violations. Therefore, in this paper, we analyze broadcast police communications (BPC) as a collaborative technology used by law enforcement agencies, specifically the Chicago Police Department (CPD). Our goal is to understand if this medium of communication (a) reflects previously observed racial disparities in CPD policing behavior \cite{noauthor_state_2019} and, related, (b) catalyzes new kinds of disparities via privacy breaches arising from everyday use of two-way radio transmissions to exchange sensitive information \cite{vargas2019digital}. Previous research confirms the external validity of using BPC in this manner.
\cite{wells1997research}.

In Chicago, use of radio technology for policing can be traced to a system purchased by a local newspaper for police use via a commercial radio station in the 1920s \cite{haller_police_1970}. By 1931, this system had been replaced with one managed by the Chicago Police Department  \cite{citizens_police_committee_chicago_1931}. A report from the time indicated radio broadcasts were already ``an integral part" of policing which ``exercises an important influence in the expeditious handling of all complaints, and in the administrative control over their investigation" \cite{citizens_police_committee_chicago_1931}. In the 1960s, Chicago's radio dispatch system was significantly upgraded to handle an increased volume of service calls \cite{miller1962chicago}. The City of Chicago's Office of Emergency Management and Communication now uses 13 dispatch zones to coordinate CPD activity in 22 police districts consisting of 277 police beats, with one dispatcher responsible for coordinating activity in 12-30 police beats each work shift. 

Such technologies (like BPC) in policing encompass a dual nature as both a sociotechnical system and a collaborative communicative system. Collaborative communicative systems function as tools facilitating effective teamwork and communication, irrespective of geographical location \cite{chand2019advanced}. On the other hand, sociotechnical systems consist of interconnected social and technical components that collectively influence goal-oriented behavior \cite{cooper1971sociotechnical, venkit2023sentiment, narayanan2023towards, venkit2024confidently} 
actors participating in their operations and technological tools used by these actors \cite{gautam2024melting}. In the case of BPC, this impact is particularly pronounced as it directly influences justice and policing in society. By conducting an in-depth examination of this collaborative policing technology our study seeks to provide novel insights into potential disparities and privacy breaches in law enforcement communication. This investigation contributes to a broader understanding of communication systems within the context of public safety and establishes a foundation for addressing challenges and enhancing the efficiency and transparency of law enforcement communication.

\section{Related Works}

The importance of information provided by police dispatchers to law enforcement officers is hard to overstate. From a privacy risk perspective, preliminary BPC research suggests that not only do police personnel communicate names and locations as a routine part of their duties, but also this exposure of personally identifiable information (PII) may be disproportionately affecting communities of color \cite{vargas2019digital}. This `digital vulnerability' identifies a potential new dimension of racial disparities in policing. This and other aspects of BPC merit further scrutiny to identify patterns in police behavior that may not be known, could plausibly pose risks to officers or members of the public, or otherwise inform relevant policy debates about the role of policing~\cite{gillooly_how_2024}.

Law enforcement agencies and others have warned that eavesdropping by `criminals' presents novel threats to officers and members of the public \cite{vargas2019digital, vargas2016wounded, mcac2011}. 
Works in privacy and surveillance have also raised various issues of surveillance in a social setting. \citet{staples1997culture} introduces the notion of the \textit{watchful gaze} to explore surveillance dynamics in various settings such as residences, and communities. Considering this, it becomes crucial to examine whether BPC serves as an intentional or unintentional means of surveillance and whether it disproportionately impacts specific populations. Furthermore, \citet{lyon2003surveillance, lyon2007surveillance} emphasizes the potential harms associated with the dissemination of individuals' personal identifiable information (PII), leading to social stratification and other social harms of an individual's privacy.

Prior research confirmed the validity of BPC for observing police behavior by matching radio transmissions and their content to relevant police records \cite{wells1997research}. However, this work concluded analysis of BPC was too resource intensive for research use. Given advances in technology, specifically fields like machine learning and speech-based applications of neural networks, we may now reconsider its viability. Specifically, studies of naturalistic speech corpora suggest multiple pathways for large scale extraction of content from these radio transmissions \cite{clifton2020100, maynard2021natural, bonin2014context}. Descriptive analysis of BPC-derived text helps demonstrate that performing this difficult task is informative for relevant research questions. 

Descriptive and lexical analysis of BPC text is salient given deficiencies in existing methodologies developed to, for example, identify the presence of racial disparities in policing \cite{neil2019methodological, prabhakaran2018detecting, voigt2017language, rho_escalated_2023}. It also provides an opportunity to demonstrate the relative value of BPC versus more difficult-to-obtain data, such as recordings from body-worn cameras \cite{rho2023escalated, voigt2017language}. While these restricted data sources capture interactions between officers and citizens, BPC enables a broader analysis of police systems since it captures the interactions between officers and 
dispatchers. 
This latter feature indicates BPC is a critical tool for officer/public safety. For example, the act of turning off one's radio places officers in danger by inhibiting their ability to ask for support, while the act of turning off one's body-worn camera does not have a similar real-time effect on officer safety. While radios are not continuously transmitting audio, their use and the content of radio transmissions represent information about how policing professionals interact during the lead-up to (and during critical moments of) an encounter which differs in nature from body-worn cameras since BPC is actively used to coordinate police activity while body-worn cameras passively record events for later review. These differences make BPC a fertile source of data for studying intra-organizational processes linkable to policing outcomes.

Consequently, in this era of rapid technological advancements, particularly in NLP and large language models (LLM), it is necessary to contemplate the implications of these innovations on personal information security, especially in accessible conversations or texts. A concerning development is the rise of privacy-invasive chatbots, adept at extracting sensitive information from diverse textual sources \cite{carlini2021extracting, kim2023propile, lukas2023analyzing, ippolito2022preventing}. As we delve into comprehending this challenge, there is a notable surge in research dedicated to unraveling privacy vulnerabilities associated via use of 
current technological breakthroughs like LLMs.


Advancing our understanding of the personal information identification capabilities of LLMs, \citet{staab2023beyond} undertakes a comprehensive examination of pretrained LLMs. Their study reveals that these models can automatically infer diverse personal attributes from extensive collections of unstructured text with an 85\% top-1 and 95.8\% top-3 accuracy. Importantly, this inference capability, coupled with the widespread adoption of LLMs, significantly reduces the costs associated with privacy-infringing inferences. This, in turn, enables malicious users to scale their operations beyond what was previously feasible with expensive human profilers \cite{staab2023beyond}.

From the research on BPC and privacy vulnerability, a noticeable gap exists in the literature dedicated to understanding this technology. No prior studies known to the authors have undertaken a comprehensive analysis of BPC to unravel its multifaceted features and assess the presence of privacy vulnerabilities concerning individuals who interact with these systems. This gap has gained particular importance in the recent era marked by the advent of large language models trained on extensive datasets, exhibiting human-level accuracy in tasks.

This gap is also significant given efforts to use historic data associated with NASA's Apollo program to advance state of the art research on naturalistic audio corpora\cite{hansen_2019_2019, hansen_fearless_2018, ziaei_speech_2014, chandra_shekar_speaker_2021,joglekar_fearless_2021}. This work spans multiple speech processing tasks and provides useful direction for study of BPC, a sociotechnical system in current widespread use. The complexity of radio communication during the Apollo program--specifically use of multiple radio frequencies, hundreds of speakers, and presence of audio noise--also reflects the complexity faced by large police systems across the world. To this end, our work advances and further establishes a nascent yet growing effort to tackle analysis of speech in natural settings where effective coordination requires transmission of accurate and timely information. Conversely, analysis of Apollo program radio communications does not pose the same privacy concerns that BPC do. Notably, efforts to analyze radio transmissions used for air traffic control represent a middle ground in this emerging space of speech-based research \cite{nguyen_n-best_2016, pellegrini_airbus_2019, badrinath_automatic_2021, guo_context-aware_2021, lin_spoken_2021, ohneiser_prediction_2021, zuluaga-gomez_bertraffic_2022}. This domain exhibits similar complexity in a contemporary sociotechnical system in widespread use but with fewer inherent privacy concerns (i.e., air traffic control audio is not used to routinely describe individuals).

The emergence of such technology necessitates heightened awareness regarding the potential misuse \cite{dev2021measures, venkit2023nationality} that may give rise to privacy concerns within datasets like BPC. This potential for misuse is expected to increase as the performance of speech recognition models on naturalistic audio improves. Our study addresses this void, representing a novel attempt to integrate an exploration of BPC's characteristics with an examination of the relationship 
between race/ethnicity and privacy vulnerability
. This study contributes to a deeper understanding of the impact of technological advancements in the present era, emphasizing the imperative of vigilance in the face of potential privacy challenges inherent in collaborative technologies like BPC.

\section{Methodology}

In this section, we describe the collected data for the purpose of analyzing BPC, along with the conducted studies aimed at comprehending the patterns and behaviors associated with BPC, and investigating the potential existence of privacy-related disparities and technological attention. Furthermore, we will describe the lexical and qualitative coding methodologies employed in the exploration of this particular mode of communication.

\subsection{Data}

The primary data used in this study draws from recordings of broadcast police communication (BPC) used to coordinate the activities of Chicago Police Department (CPD) officers. We obtained BPC audio recordings from Broadcastify.com. Data were downloaded after paying a nominal subscription fee to access six months of archived recordings. Two rounds of data collection were conducted six months apart to obtain approximately one year of data. Each file in the archive is over 30 minutes long, stored in MP3 format, and represents a continuous recording of one radio frequency. Data were made available by the site owner under a Creative Commons Attribution 3.0 United States license \cite{creative2023}.

A total of $161,550$ files were obtained, representing $80,775$ hours of audio linked to CPD activities from August 1, 2018 to July 31, 2019. For the studies here, we focus on the time frame of August 10th, 2018 from 9:00 AM to 5:00 PM. This time frame represents a typical Friday workday in Chicago during the Summer (high/low temperature of 83$^{\circ}$F/70$^{\circ}$F compared to mean high/low temperature of 83$^{\circ}$F/66$^{\circ}$F, no recorded precipitation).

\begin{figure}
\centering
\includegraphics[scale = 0.25]{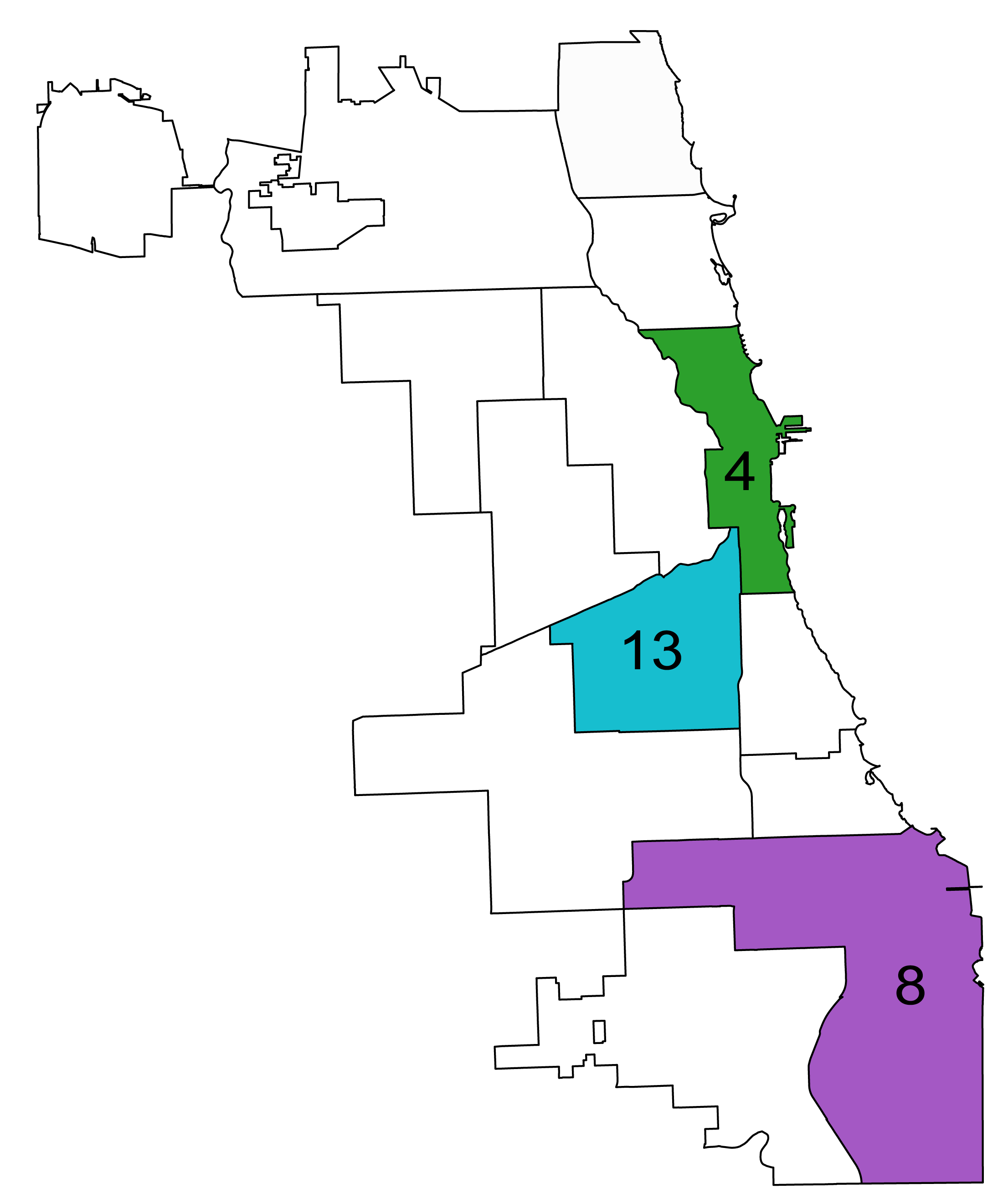}
\caption{Zone 4, Zone 8 and Zone 13 represent a majority White, Black or African American, and Hispanic group of communities, respectively. White regions denote other dispatch zones in the city of Chicago.} \label{zones}
\end{figure}

\begin{figure}
\centering
\includegraphics[scale = 0.3]{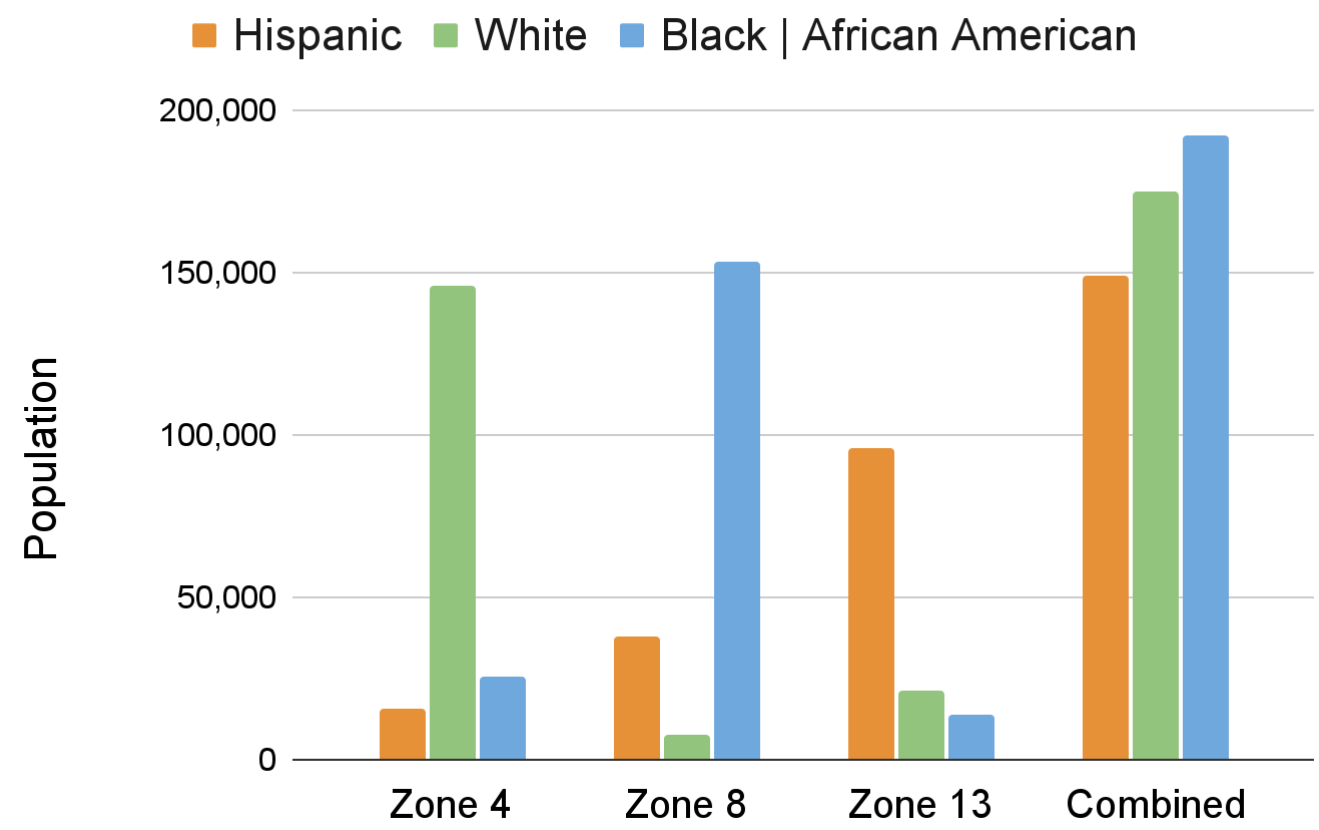}
\caption{Population distribution of Hispanic, White and Black or African American population across Zone 4, Zone 8 and Zone 13.} \label{census}
\end{figure}

Following selection criteria used by Vargas et al. \cite{vargas2019digital} to analyze police-dispatcher communication, we study transcripts of BPC from three \textit{dispatch zones}. A dispatch zone is a geographic area used to coordinate CPD activity, and the selected zones--Zone 4, Zone 8, and Zone 13--in Fig. \ref{zones} serve predominately White, Black or African American, and Hispanic residents, respectively. Population distribution is shown in Fig. \ref{census} and is based on data from the 2020 decennial census. 

\begin{figure*}
\centering
\includegraphics[scale = 0.23]{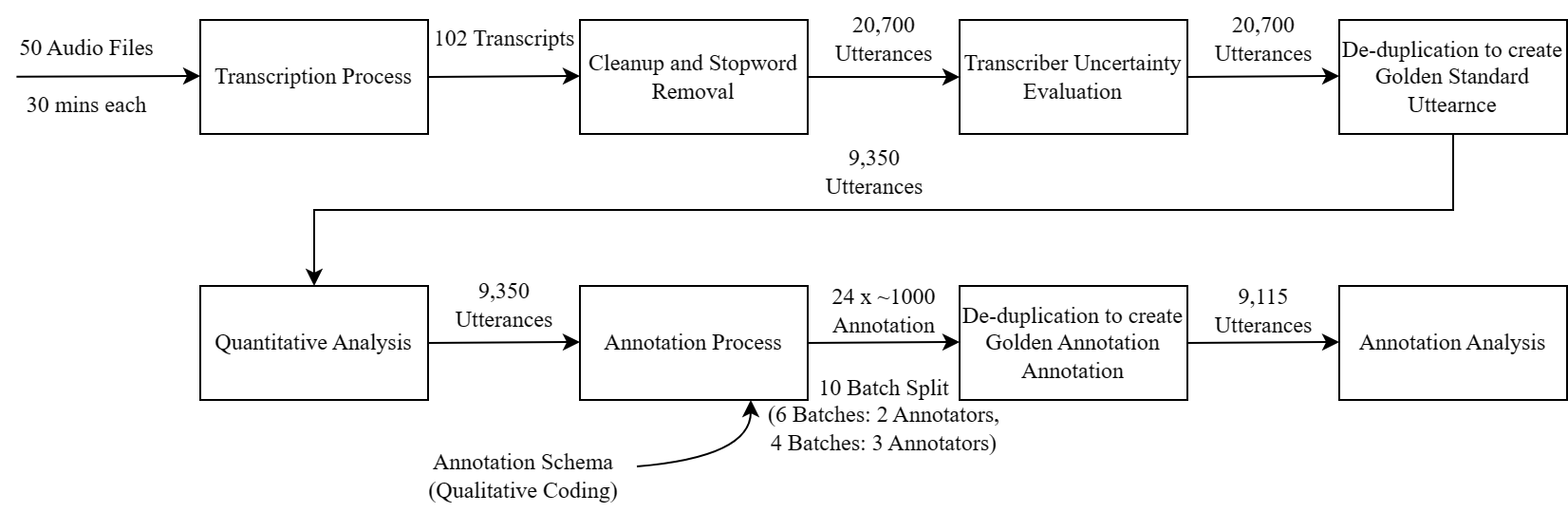}
\caption{Flowchart depicting each procedure to obtain the required data for analysis.} \label{fig:dataflowchart}
\end{figure*}

We obtain a total of \textbf{9,115} utterances for analysis after the transcription process, with the following utterance division for each zone: \textbf{Zone 4 = 3,333, Zone 8 = 3,921 and Zone 13 = 1,861}. We use this corpus to perform our descriptive analysis and thematic qualitative coding. Figure \ref{fig:dataflowchart} depicts the data collection process in detail.

\subsection{Data quality}

Transcription was performed at University of Chicago, with each transmission transcribed by two transcribers to address difficulties in understanding transmissions and variability in audio quality. Transcribers were instructed 
to indicate passages, where they were 50\% to 90\% confident in the content, with a similar strategy used when they had little to no confidence they understood what was said. Audio with little uncertainty in content (i.e., >90\% confidence) was entered as-is. While ground truth transcriptions do not exist for these data, we used word error rate, a common metric for assessing the performance of speech recognition systems, to assess inter-transcriber agreement. 

Word error rate (WER) measures the number of words that must be inserted, deleted, and substituted in hypothesized text to match ground truth text. The lower bound of WER is 0, indicating hypothesized text matches ground truth text, but the upper bound of WER can easily exceed 1 if hypothesized text includes more words than its corresponding ground truth text.
Two-word error rates were calculated for each pair of transcribers, using one transcriber and then the other as the ``ground truth'' transcription for each 30-minute file. All audio was hand-labeled by two transcribers to generate a gold standard in lieu of ground truth, which allows us to identify patterns in relative agreement between transcribers. This approach identified one transcriber whose WER (i.e., level of disagreement with all other transcribers who annotated the same audio) was typically much higher than other transcribers. Accordingly, their transcripts were excluded from analysis. Otherwise, the median inter-transcriber WER was 0.23 for the $n=370$ pairwise comparisons made. This is lower than the word error rate achieved by a professional transcription service (WER = $0.30$) on a test set of audio clips chosen for their high quality (i.e., low noise, clear speech, etc.).

To generate gold standard transcripts, we developed a metric based on transcriber certainty about BPC content to assess transcript quality. 
Text from the transcriber exhibiting 
the most certainty 
in their transcript was 
selected to remove redundancy in the transcription. The score is intended to penalize transcripts based on amount of uncertainty 
and reward them based on the amount of content transcribed. This metric facilitated consistent selection criteria for identifying which transcript to use given redundant transcriptions of the same audio file. For robustness, we analyzed text from transcripts with the highest and lowest uncertainty to increase our confidence in the transcription process: Our results show similar patterns between the two set of transcriptions, suggesting that non-LEO transcribers produced substantively consistent transcriptions and the choice of transcripts based on uncertainty did not affect the patterns observed.

We did not attempt to validate our analysis of hand-labeled text using a larger sample of machine-labeled text from our BPC archive, as text inferred from a state of the art, noise robust speech recognition model (i.e., Whisper large-v3 \cite{radford_robust_2022}) exhibited a WER of $0.81$ when compared to the gold standard obtained using the process above. This level of inaccuracy/disagreement far exceeds inter-transcriber WER values and is comparable to the WER values generated by the annotator whose transcripts were excluded from analysis due to especially low quality.

\subsection{Study 1: Validate presence of racial/ethnic disparities in BPC}

The content of BPC is important given its purpose of coordinating police activity. The functional use of this language has immediate implications for what aspects of BPC should be studied. Since radio frequencies must be kept `open' in case officers require critical, immediate assistance, it is reasonable to assume transmissions are as short as possible to ensure channels remain available for urgent communication when needed most. 
This use case suggests lexical analysis is vital since each word may be essential for directing police activity, and thus disproportionate mention of any group would indicate a disparity in police attention.
\subsubsection{Comparative analysis of utterance length} To validate the relative importance of lexical analysis for identifying racial/ethnic disparities, we compare BPC utterance lengths to similar contexts. This is useful since BPC transmission length has not been analyzed in this way so we cannot make the assumption transmissions are particularly short (i.e., information dense) relative to other speech contexts. We compare use of BPC to air traffic control and phone calls and (1) confirm BPC transmissions are shorter than functionally similar air traffic control transmissions and (2) find the word count distribution of BPC utterances is more similar to phone calls than air traffic control communications, suggesting mentions of race/ethnicity are important given BPC radio transmissions often contain few words.

While there is little research on BPC as a collaborative communication system, air traffic control represents a similar sociotechnical system \cite{sewell_forgotten_1984} with substantially more research on how such radio-based systems function \cite{fowler_air_1980, manning_using_2002, skaltsas_analysis_2013, nguyen_using_2016, nigmatulina_two-step_2022}. We identify three main points of similarity: (a) use of call signs and call-and-response identification processes; (b) a one-to-many functional relationship between a central coordinator and multiple actors seeking information and direction; (c) the high stakes of system failures. In light of these functional similarities, we use the ATC0 corpus of air traffic control communications \cite{godfrey1994air}, composed of radio transmissions at three major American airports, to contextualize BPC transmission length. Among available air traffic control corpora, ATC0 best fits BPC given it is largely composed of native English speakers and contains 70 hours of annotated audio \cite{pellegrini_airbus_2019}. 
We also compare BPC transmission length against the Switchboard corpus due to its intuitive context and use case (i.e., phone calls), which reflect similar back and forth communications between individuals as found in BPC. 

\begin{figure*}
\centering
\includegraphics[scale = 0.32]{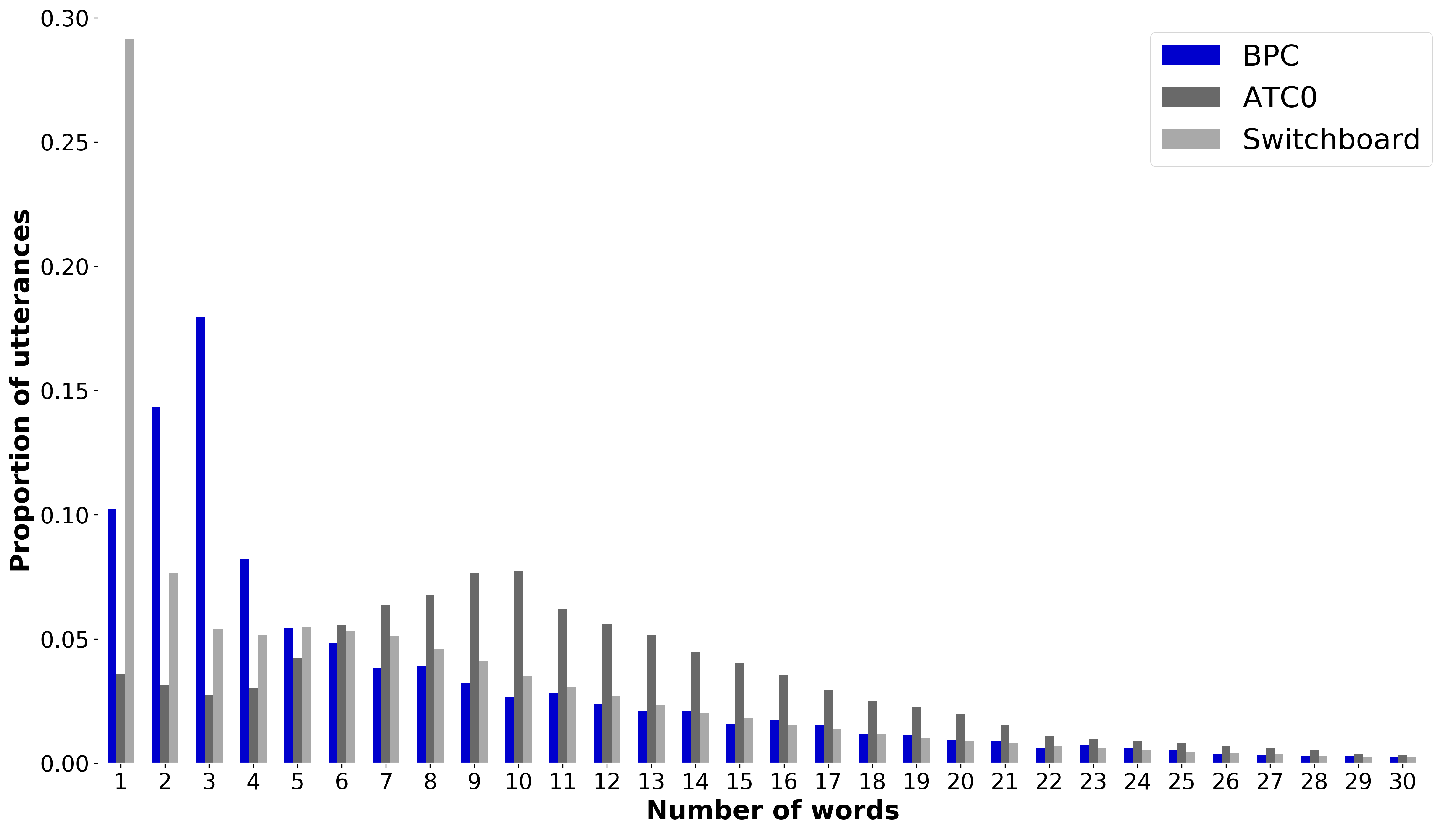}
\caption{Relative distribution of word frequencies for utterances in the BPC, ATC0, and Switchboard datasets} \label{fig2}
\end{figure*}

Analysis shows BPC utterances are more akin in length to those in phone conversations than to communications between air traffic controllers and airline pilots, despite the similar functional context of radio use in policing and commercial flight. Specifically, BPC utterances exhibit a long, thin tail to the right, while ATC0 and Switchboard utterances exhibit similar distributions (Fig. \ref{fig2}). Related, the modal ATC0 utterance is 10 words long, the modal BPC utterance is 3 words long, and the modal Switchboard utterance is 1 word long (e.g., ``Hello"). To quantify this comparison, we calculated Kullback–Leibler divergence for the discrete probability distributions constituted by utterance length frequency counts, using BPC as the ``true'' distribution such that a value of 0 indicates a distribution identical to BPC. Values greater than 0 indicate asymmetric distributional differences, with higher values indicating greater differences. We confirmed differences are greater for ATC0 ($D_{KL} = 0.50$) than for Switchboard ($D_{KL} = 0.22$) despite functional similarities between use of BPC and ATC.

\subsubsection{Disproportionate police attention} Given the succinct nature of BPC, analyzing its content is an important next step for validating the presence of racial disparities in police attention. Due to the prevalence of police jargon in BPC transmissions, including domain-specific syntax to describe personal characteristics (e.g., ``male black" for a Black or African American male), we employ bigram analysis to explore this domain-specific language. While unigram frequency provides informative insights, policing language often consists of coded language that comprises two or more words. However, for example, frequent use of numerals in BPC transmissions renders them ambiguous and challenging to analyze without additional context. 
Further, the term ``black" may reflect a person description but may also reference an object and we cannot know the referant without additional linguistic context. Notably, words associated with race/ethnicity always appear as part of bigrams with gender terms. Therefore, we focus on bigram frequencies to overcome these challenges and filter out bigrams containing stopwords to emphasize 
context. Our stopword list is derived from the NLTK library\footnote{https://www.nltk.org/nltk\_data/} excluding words associated with interrogatives and gender-related terms, which are meaningful for our analysis.

\begin{table*}
\centering
\footnotesize
\begin{tabular}{|r|r|r|r|r|}
\hline
& \textbf{All Zones} &  \textbf{Zone 4} & \textbf{Zone 8} & \textbf{Zone 13} \\
\hline
1 & go ahead [133]&  event number [52]& traffic stop [70]& go ahead [38]\\
\hline
2 & traffic stop [97]& \textbf{male black} [41]& go ahead [65]& last seen [27]\\
\hline
3 & \textbf{male black} [88] & la selle [36]& unit coming [41]& 104 sir [21]\\
\hline
4 & event number [85]&  go ahead [32]& \textbf{male black} [33]& alright 104 [18]\\
\hline
5 & alright 104 [59]&  north michigan [28] & alright 104 [23] & \textbf{male white} [17]\\
\hline
6 & unit coming [53]& 181 charlie [21]& event number [23] & \textbf{male black} [17]\\
\hline
7 & i'm gonna [50]&  alright 104 [20]& go head [20] & sir go [14]\\
\hline
8 & dont know [42]&  oh north [17]& crime scene [20] & gray shirt [13]\\
\hline
9 & show us [38]&  show us [17]& seventy sixth [18] & traffic stop [12]\\
\hline
10 & i'm sorry [38]&  i'm gonna [17]& cottage grove [18] & event number [12]\\
\hline
\end{tabular}
\caption{Top 10 bigrams for utterances from Zones 4, 8, 13 and from all zones are shown in this table. The frequency of each bigram is shown in `[ ]' adjacent to the word. Terms related to race are shown in bold to highlight their presence.}\label{tab1}
\end{table*}

Our analysis of bigram 
, shown in Table \ref{tab1}, detected multiple terms related to race/ethnicity and gender in addition to location-based descriptions, event numbers, and other procedural terms, like ``alright 104'', used to coordinate police activity. 
We find the most common race-related bigrams across all three zones are ``male black'' and ``female black'', and that 
the bigram ``male black" frequently appears in all zones, including in Zone 4, a predominantly white neighborhood.

\begin{figure}
\centering
\includegraphics[scale = 0.4]{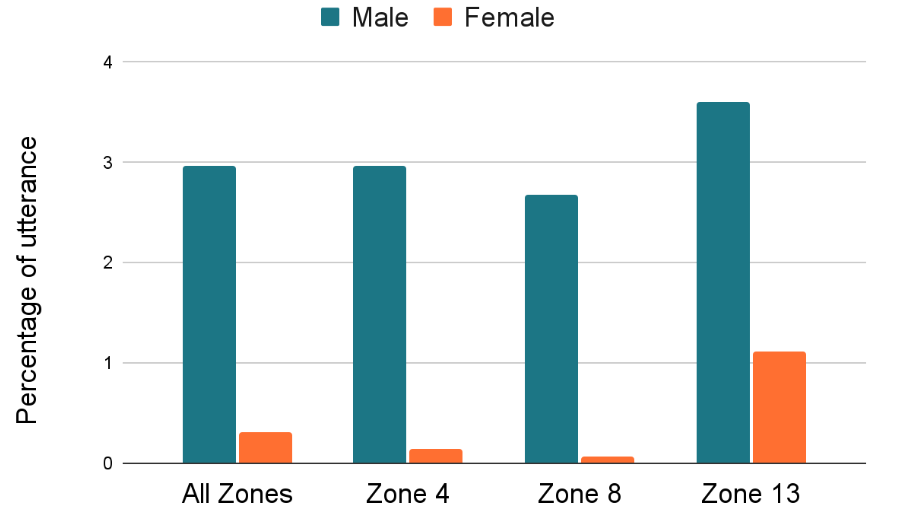}
\caption{Percentage of utterances containing terms related to gender in BPC.} \label{gender}
\end{figure}

Notably, male-gendered terms appear nine times more frequently than female-gendered terms in these conversations (see Fig \ref{gender}). We observe this difference in the presence of male and female-gendered terms across all three zones, 
though Zone 13 displays a larger percentage (almost four times) of statements referring to the female-gendered terms than other zones. In our analysis, we include all forms of gender-related terms identified in the utterances, including plural forms. Our findings reveal that, irrespective of race/ethnicity, police give disproportionate attention to males compared to females vis-a-vis BPC mentions of gender. 

\begin{table*}[]
\centering
\footnotesize
\begin{tabular}{|c|ccc|ccc|}
\hline
\textbf{} & \multicolumn{3}{c|}{\textbf{Male}} & \multicolumn{3}{c|}{\textbf{Female}} \\ \hline
\textbf{} & \multicolumn{1}{c|}{\textbf{Black}} & \multicolumn{1}{c|}{\textbf{Hispanic}} & \textbf{White} & \multicolumn{1}{c|}{\textbf{Black}} & \multicolumn{1}{c|}{\textbf{Hispanic}} & \textbf{White} \\ \hline
\textbf{All Zones} & \multicolumn{1}{c|}{1.19} & \multicolumn{1}{c|}{0.12} & 0.31 & \multicolumn{1}{c|}{0.25} & \multicolumn{1}{c|}{0.03} & 0.02 \\ \hline
\textbf{Zone 4} & \multicolumn{1}{c|}{1.22} & \multicolumn{1}{c|}{0.13} & 0.15 & \multicolumn{1}{c|}{0.18} & \multicolumn{1}{c|}{$\leq$ 0.01} & 0.05 \\ \hline
\textbf{Zone 8} & \multicolumn{1}{c|}{1.16} & \multicolumn{1}{c|}{0.02} & 0.07 & \multicolumn{1}{c|}{0.27} & \multicolumn{1}{c|}{$\leq$ 0.01} & $\leq$ 0.01 \\ \hline
\textbf{Zone 13} & \multicolumn{1}{c|}{1.18} & \multicolumn{1}{c|}{0.32} & 1.12 & \multicolumn{1}{c|}{0.37} & \multicolumn{1}{c|}{0.16} & $\leq$ 0.01 \\ \hline
\end{tabular}
\caption{Percentage of utterance containing gender-demography bigrams categorized by zones.}\label{tab2}
\end{table*}

Finally, we quantify the relative joint prevalance of bigrams indicating gender and race/ethnicity to characterize disparities in police attention independently observed in each. Table \ref{tab2} shows strong within-gender and between-gender differences in police attention, where males receive more attention than females, in general, but Black males receive the most attention of all groups. Multi-way analysis of variance with percent of utterances as outcome suggests statistically significant differences by gender ($F=12.18$, $df=1$, $p=0.005$) and race/ethnicity ($F=7.68$, $df=2$, $p=0.007$) but not zone ($F=1.54$, $df=2$, $p=0.254$), though sample size (n = 18) limits the conclusions we can draw from this analysis.

\subsection{Study 2: Describe patterns in privacy vulnerability within BPC}

While the previous study presents lexical analysis that validates the presence of racial/ethnic disparities in BPC content, this study describes separate qualitative analysis of BPC content to identify patterns in privacy vulnerability. 
To achieve this goal, we conducted thematic qualitative coding that involved human evaluation to create an exhaustive taxonomy based on the speech act of the text. We identify six classes of speech acts and find one class tends to contain sensitive information more often than the other five. To understand the nature of privacy vulnerablities, we also characterize the prevalence of sensitive information across these classes of speech acts.

Our thematic qualitative coding \cite{forman2007qualitative} involved an iterative reading process, where we generated and validated an exhaustive list of taxonomies to differentiate the encountered utterances based on their apparent purpose. All authors performed close reading of a random selection of 2000 utterances from the combined three zones as part of the coding process, which was divided into four phases: \textit{formalization with data, generation of initial codes, searching and reviewing themes,} and \textit{defining the taxonomy for annotating the utterances}. 

Our qualitative analysis yields six groups that classify utterances based on their apparent purpose: \textit{Event Information Transmission, Procedural Transmission, Liminal Transmission, Miscellaneous Policing Transmission, Casual Transmission,} and \textit{Unclear Intend Transmissions}. The description of each group is mentioned in Table \ref{table:definition}. We evaluated the reliability of our categories using an annotation agreement test, following the guidelines provided by \citet{mcdonald2019reliability}, where four annotators who were not involved in the coding process annotated 1000 utterances. The annotation process resulted in \textbf{71\%} agreement amongst the annotators, with a Fleiss Kappa value of \textbf{0.706}, indicating substantial agreement regarding the categorization of utterances based on their apparent purpose.

In addition, we defined four other features that were used to annotate BPC for sensitive information: 
Protected Health Information (PHI), gender, race/ethnicity, and age. PHI is the health data created, received, stored, or transmitted by HIPAA-covered entities and their business associates in relation to healthcare services\footnote{https://www.hhs.gov/hipaa/for-professionals/privacy/laws-regulations/index.html}. In this study, PHI is defined as the information that can be used to identify an individual directly or indirectly (in conjunction with other data elements). Our PHI taxonomy consisted of 13 labels identifying various data elements that can be used to identify an individual directly or indirectly, which was taken based on the HIPAA-compliant PHI list of identifiers \cite{ocr2023phi}. For gender, race/ethnicity, and age parameters, we provided guidelines that required annotators to quote these sociodemographic details as found in the utterance.

\begin{table*}[]
\centering
\footnotesize
\begin{tabular}{|c|l|l|}
\hline
\textbf{Class Name} & \multicolumn{1}{c|}{\textbf{Definitions}} & \multicolumn{1}{c|}{\textbf{Examples}} \\ \hline
\begin{tabular}[c]{@{}c@{}}Event Information \\ Transmissions\\(EVENT)\end{tabular} & \begin{tabular}[c]{@{}l@{}}Transmissions that consist of reported\\ activity and follow-up information\\  or questions.\end{tabular} & \begin{tabular}[c]{@{}l@{}}1. Residential alarm \textless{}redacted\textgreater break-in\\ 2. I need an ID for 102 mission.\end{tabular} \\ \hline
\begin{tabular}[c]{@{}c@{}}Procedural\\ Transmissions\\(PROC)\end{tabular} & \begin{tabular}[c]{@{}l@{}}Transmissions that consist entirely\\ of or almost entirely of police jargon.\end{tabular} & \begin{tabular}[c]{@{}l@{}}1. 712\\ 2. 104 now 3914 on \textless{}redacted\textgreater{}\end{tabular} \\ \hline
\begin{tabular}[c]{@{}c@{}}Liminal\\ Transmissions\\(LIM)\end{tabular} & \begin{tabular}[c]{@{}l@{}}Transmissions chiefly acknowledging\\ recent requests or statements in other\\ transmissions.\end{tabular} & \begin{tabular}[c]{@{}l@{}}1. Thank you.\\ 2. Go for Robert.\end{tabular} \\ \hline
\begin{tabular}[c]{@{}c@{}}Miscellaneous Policing\\ Transmissions\\(MISC)\end{tabular} & \begin{tabular}[c]{@{}l@{}}Transmissions about policing duties\\ but are unrelated to reported activity.\end{tabular} & \begin{tabular}[c]{@{}l@{}}1. 1 2 3 4 Radio Check\\ 2. Just for your information \textless{}redacted\textgreater\\ lights blinking.\end{tabular} \\ \hline
\begin{tabular}[c]{@{}c@{}}Casual Transmission\\ (CAS)\end{tabular} & \begin{tabular}[c]{@{}l@{}}Transmissions with clear intent that\\  do not qualify as Event, Procedural,\\ Liminal or Miscellaneous.\end{tabular} & \begin{tabular}[c]{@{}l@{}}1. Morning squad. Happy Friday.\\ 2. I'm good. Just a little bit tired.\end{tabular} \\ \hline
\begin{tabular}[c]{@{}c@{}}Unclear Intent\\ Transmissions\\(UIT)\end{tabular} & \begin{tabular}[c]{@{}l@{}}Transmissions where the intent\\ is unclear.\end{tabular} & \begin{tabular}[c]{@{}l@{}}1. {[}background noise{]}\\ 2. {[}noise{]}\end{tabular} \\ \hline
\end{tabular}
\caption{Definition and examples of categories obtained through thematic qualitative coding of BPC based on their purpose.} \label{table:definition}
\end{table*}

We divided the utterances into ten batches for the annotation process, which five research assistants from major, research-I universities\footnote{Information redacted for anonymity} carried out. Each batch was annotated by at least two (6 batches) and a maximum of three (4 batches) annotators. The values for speech act were consolidated based on the majority annotation for each utterance. To resolve disagreements and arrive at a consensus on the PHI value annotation, annotators engaged in a consolidation discussion for each batch. The remaining sociodemographic parameters were left unmodified, as they had been directly annotated based on their presence in the utterance. 

\begin{figure}
\centering
\includegraphics[scale = 0.3]{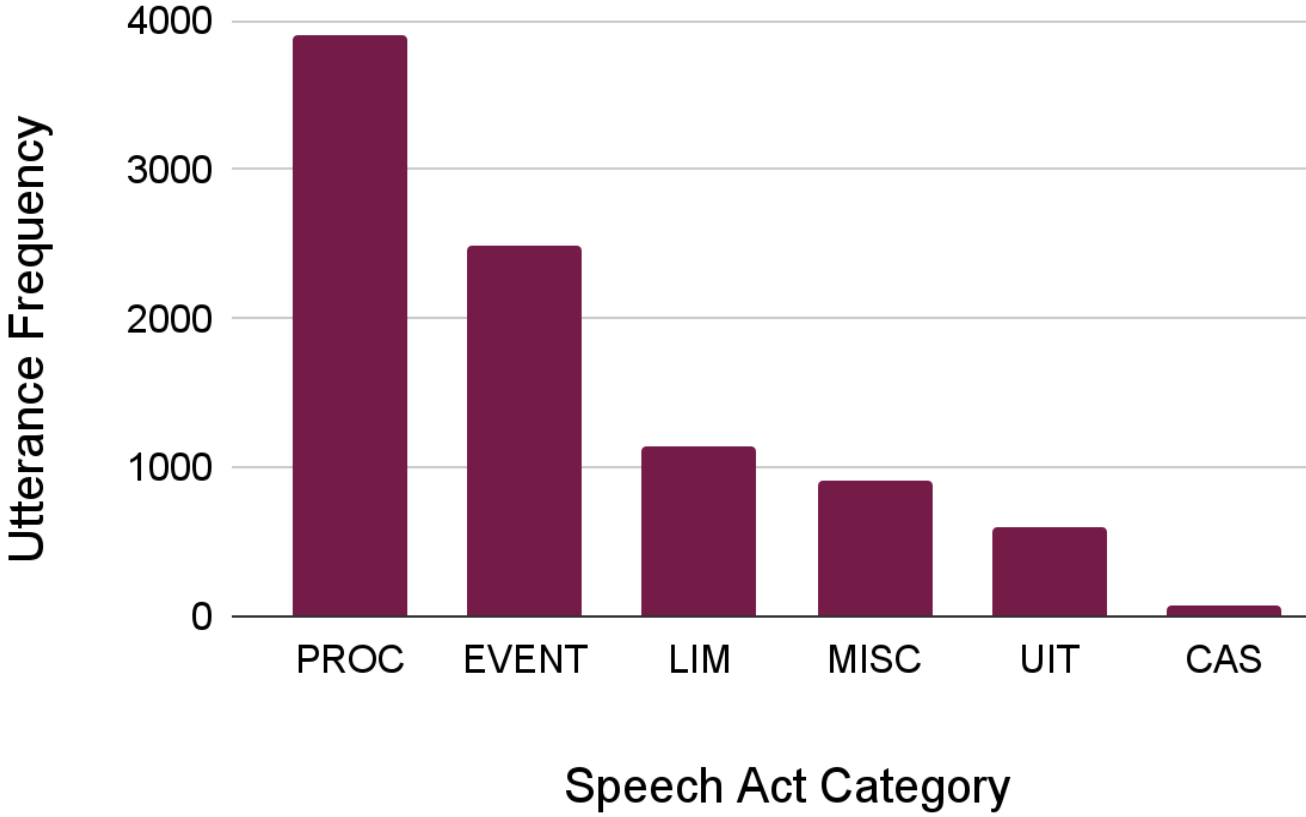}
\caption{Utterance frequency of BPC for each annotation class categorized by purpose.} \label{fig:speechact}
\end{figure}

Our findings indicate Procedural utterances are the most frequent type of communication, followed by Event Information transmissions (Fig. \ref{fig:speechact}). Human evaluation of the BPC dataset confirms that short, coded sentences are used to address critical, immediate needs. We also examine the word frequency distribution of BPC by category of speech act to identify trends. 

\begin{figure}
\centering
\includegraphics[width=8.5cm, height=5.5cm]{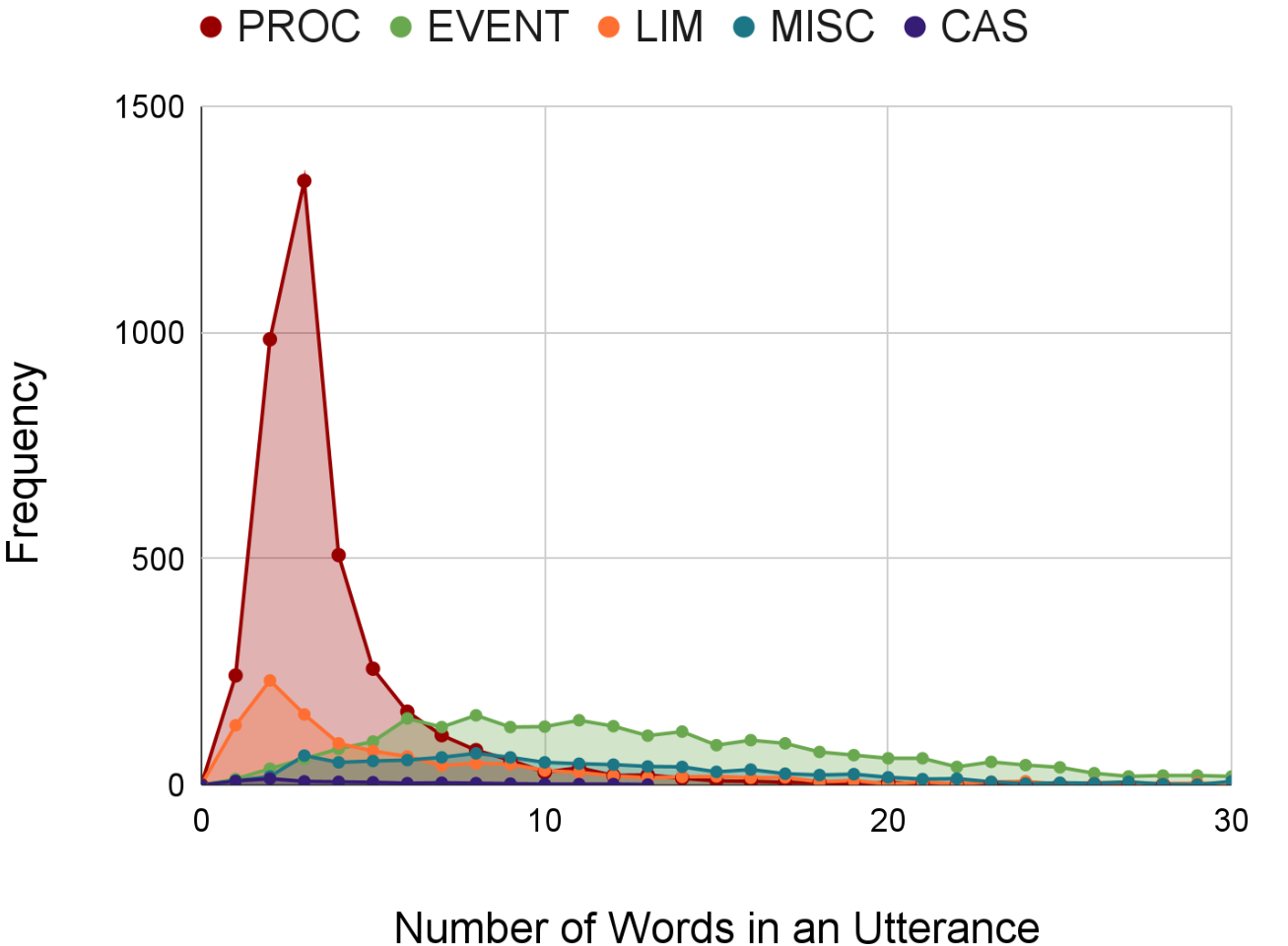}
\caption{Frequency distribution of utterance in each class categorized by their purpose. UIT class is removed as the utterances are classified as noise.} \label{fig:ClassUtterance}
\end{figure}

In Figure \ref{fig:ClassUtterance}, we observe the frequency of words in each utterance categorized according to their apparent purpose. For Procedural transmissions 
, the maximum length of an utterance is limited to 38 words, with a mode of 3 words. In contrast, Event Information transmissions can have up to 110 words, with a mode of 13 words, resulting in more descriptive utterances. We hypothesize the distribution of word frequency across the classes highlights the distinct nature of each class. 

Our annotations reveal that BPC is information-dense, based on the frequency of each class, and can provide insights into incidents even for individuals without background knowledge of the coded language used in these communications. However, this also highlights the potential for detrimental eavesdropping and misuse. In particular, Event Information transmissions contribute to the long tail seen in Fig. \ref{fig2}, indicating this type of BPC has the most potential for misuse, if only because these transmissions tend to be the longest (i.e., contain the most words/information).

\begin{figure}
\centering
\includegraphics[scale = 0.3]{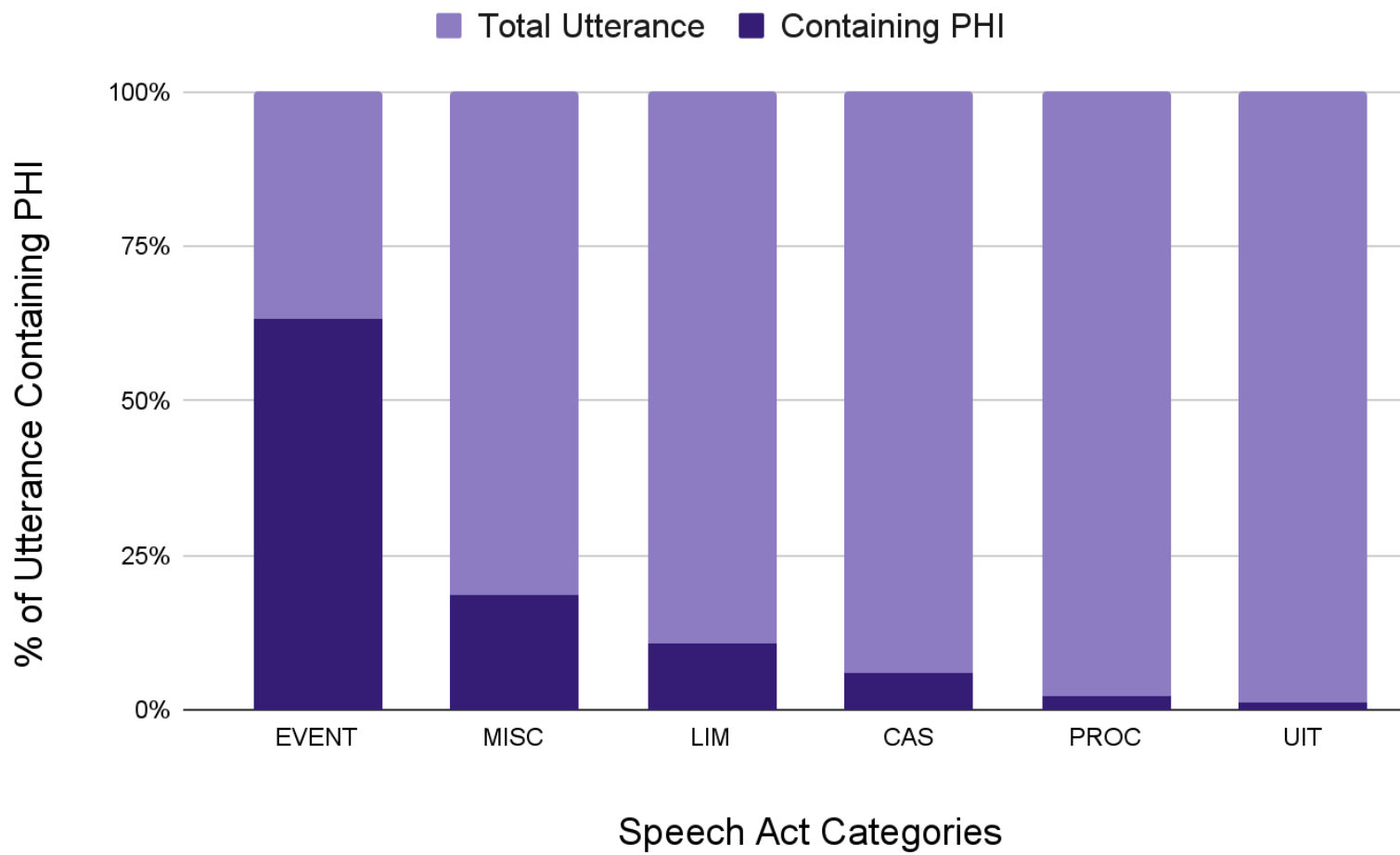}
\caption{Percentage of utterance containing PHI in a speech act category.} \label{fig:phispeech}
\end{figure}

\begin{figure}
\centering
\includegraphics[scale = 0.3]{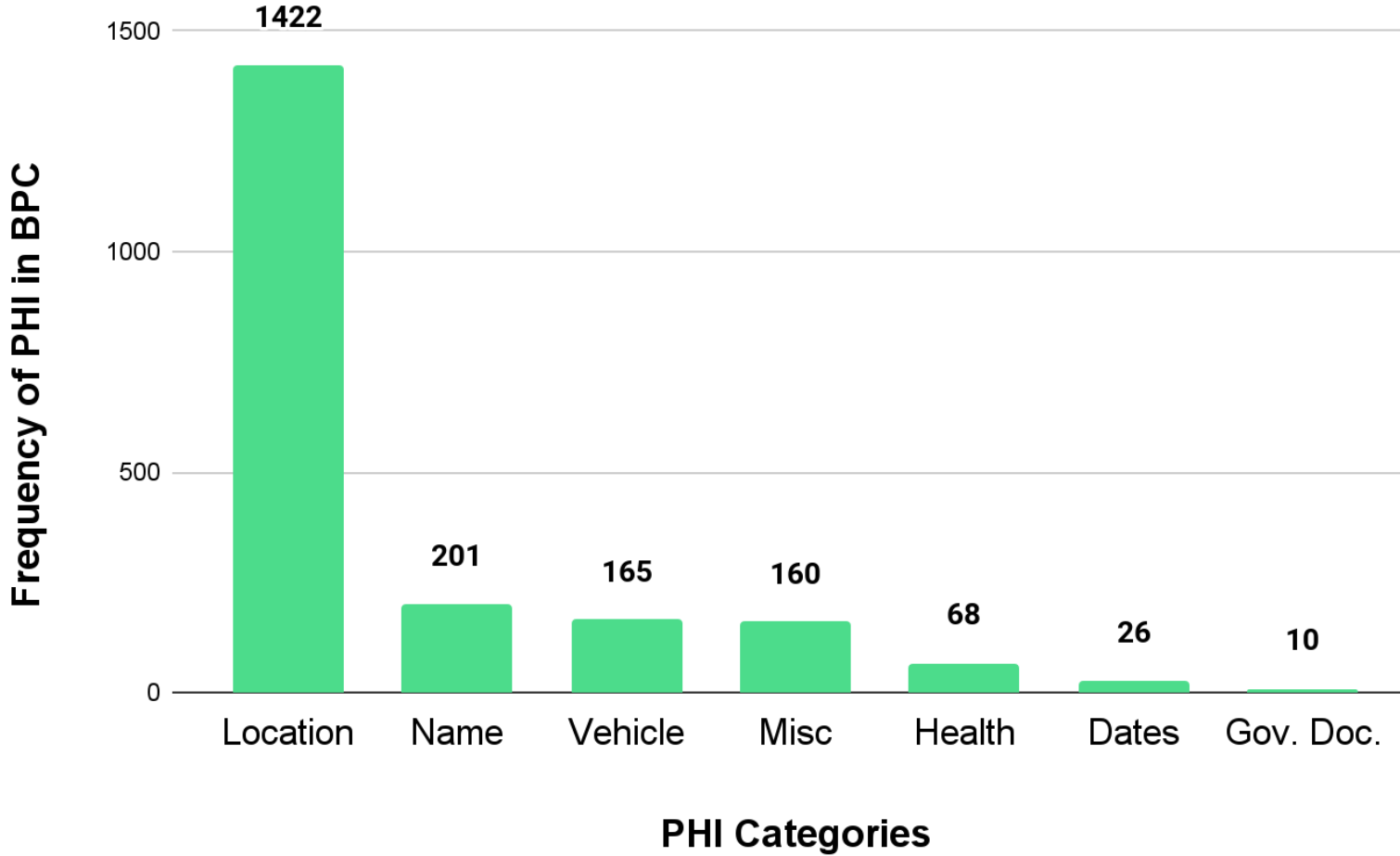}
\caption{Categories of PHI and their corresponding utterance frequency.} \label{fig:phitype}
\end{figure}

We investigate privacy vulnerablity further by evaluating the PHI parameter, identified during our annotation process. The PHI parameter indicates how BPC pose potential privacy concerns through the interaction with the selected technology. Out of 9115 utterances analyzed, \textbf{1817} contained at least one mention of PHI. We find that \textit{event information transmission} constitutes the largest number of PHI mentions (60\% of utterances) among various speech act categories, as illustrated in Fig. \ref{fig:phispeech}. We also find 13\% of the utterances with PHI contained at least two different mentions of PHIs, while 87\% contained a single mention of PHI.

Additionally, we investigated how speakers in different dispatch zones reveal PHI to understand potential differences in behavior. All three selected zones showed a similar percentage of utterances containing PHI (Zone 4: 22.7\%, Zone 8: 16.6\%, Zone 13: 21.7\%). This indicates PHI is present in BPC from all zones and demonstrates that, independent of geographic context, BPC is a medium likely to catalyze privacy vulnerability.

Fig. \ref{fig:phitype} depicts the types of PHI most commonly shared via BPC, with an individual's location and first and last name being the most frequently mentioned. Moreover, individuals' vehicular details, health information, 
and government-issued document details were also shared on BPC. We note that \textit{any combination of these PHIs can be exploited to reveal an individual's identity}, leading to further data and privacy losses \cite{schwartz2011pii, venkatadri2018privacy}. It is, therefore, crucial to understand the consequences of disclosing such data on a public channel, such as BPC. Additionally, we found that sensitive information such as an individual's license ID number and other government-issued document details were communicated openly on BPC.

\subsection{Study 3: Assess BPC-related attention disparity and privacy vulnerability across racial/ethnic groups}

\textbf{Attention disparity by gender, race/ethnicity, and age:} In this study, we explore the digital surveillance and attention demonstrated by BPC across various races and ethnicities in each zone. In Staple's \cite{staples1997culture} work, the term \textit{surveillance} is characterized as the active monitoring of individuals. It posits the notion of a `watchful gaze', which pertains to routine, localized data-gathering practices that occur within various settings, such as workplaces, schools, residences, and communities. Building upon this foundation, Lyon \cite{lyon2003surveillance, lyon2007surveillance} expands the concept to encompass the methodical accumulation, scrutiny, and distribution of personal information with the aim of exerting control, providing guidance, or safeguarding certain individuals or cohorts. Notably, implementing surveillance tactics targeting specific groups has been shown to amplify attention toward them and potentially result in social stratification \textit{absent their explicit consent}. Based on these work, we define \textit{technological attention} as the \textit{watchful gaze of a technology towards a specific demographic group, with or without their consent}. To analyze \textit{technological attention}, we examine the frequency of utterances that reference a particular race/ethnicity group and compare it to the overall population of that race/ethnicity. Any discrepancies in this distribution may suggest a possible \textit{disproportionality} in the frequency of mentions of that race within the zone, demonstrating greater technological attention towards that group.

Along with race/ethnicity, we examined gender and age categories defined during the annotation process. Our analysis revealed that the utterances primarily mention female and male-related terms, with only White, Black or African American, Hispanic, Middle Eastern, and Light Skin racial-ethnicity groups present. Excluding Middle Eastern and Light Skin groups, which were mentioned in just one utterance, Table \ref{table:race} displays the number of utterances mentioning both race and gender terms either as a bigram or in the same utterance. Our findings indicate that the majority of the utterances, that mention a sociodemographic indicator, referred to male gender terms ($\sim$68\%) and Black race/ethnicity terms ($\sim$69\%). Consistent with our descriptive analysis using bigrams, we observe that African American terms mentioning male gender were the most frequent group among all combinations of gender and race/ethnicity. The categorization of race/ethnicity references in relation to individual social groups, along with their respective population distributions within the specified zone, is illustrated in Fig. \ref{fig:racementions}. This depiction serves to underscore the observed attention disparity within the context of BPC.

To test if the distribution of mentions by gender and race/ethnicity found in Table \ref{table:race} represents a statistically significant departure from random chance (i.e., one definition of equitable attention), we consider a null baseline where mentions are expected to be equal in number for all groups. In this context, we reject the null hypothesis that mentions are uniformly distributed across the six combinations of gender and race/ethnicity reported here (${\chi}^2(5,N=192)=202.9$, $p<0.001$). However, this test is unrealistic since geographic context, like the demographic composition of local residents, does not contribute to the expected frequency of each cell. To this end, we also consider a population-based baseline where mentions are expected to be proportional to the relative prevalence of each group within the dispatch zones under study. 
Using this baseline, we again find differences to be statistically significant (${\chi}^2(5,N=192)=187.8$, $p=<0.001$). Results indicate disproportionate police attention of male and/or Black or African American residents 
, as it is clear a disparity in police attention exists within these six demographic groups since male and/or Black or African American individuals are mentioned much more often than other groups.

\begin{table}[]
\centering
\footnotesize
\begin{tabular}{|c|c|c|}
\hline
\textbf{Gender} & \textbf{Race/Ethnicity} & \textbf{Utterance Freq.} \\ \hline
Female & Black & 31 \\ \hline
Female & Hispanic & 8 \\ \hline
Female & White & 8 \\ \hline
Female & Unmentioned & 163 \\ \hline
Male & Black & 102 \\ \hline
Male & Hispanic & 11 \\ \hline
Male & White & 32 \\ \hline
Male & Unmentioned & 318 \\ \hline
\end{tabular}
\caption{Frequency of utterance with mentions of race/ethnicity and gender.} \label{table:race}
\end{table}

\begin{figure}
\centering
\includegraphics[scale= 0.3]{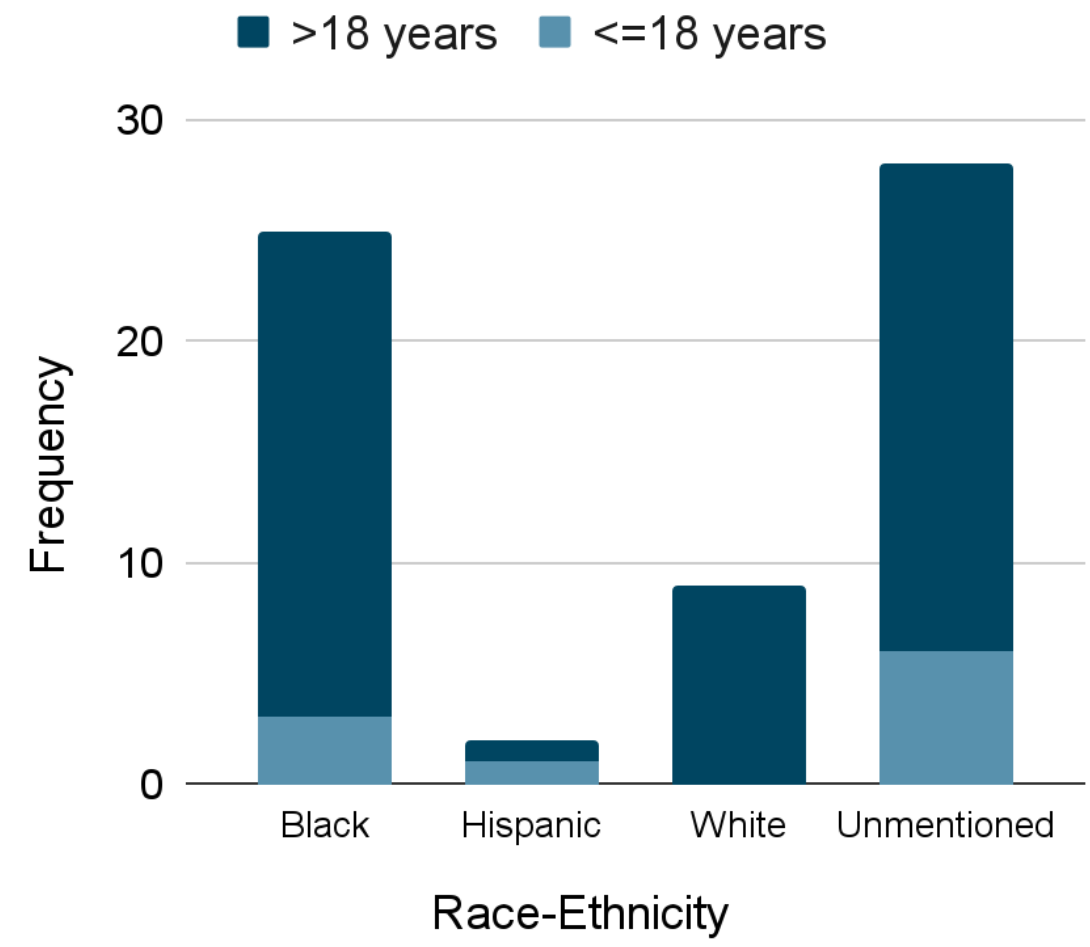}
\caption{Frequency of utterances (not) mentioning developmental status.} \label{fig:agedist}
\end{figure}

\begin{figure}
\centering
\includegraphics[scale = 0.3]{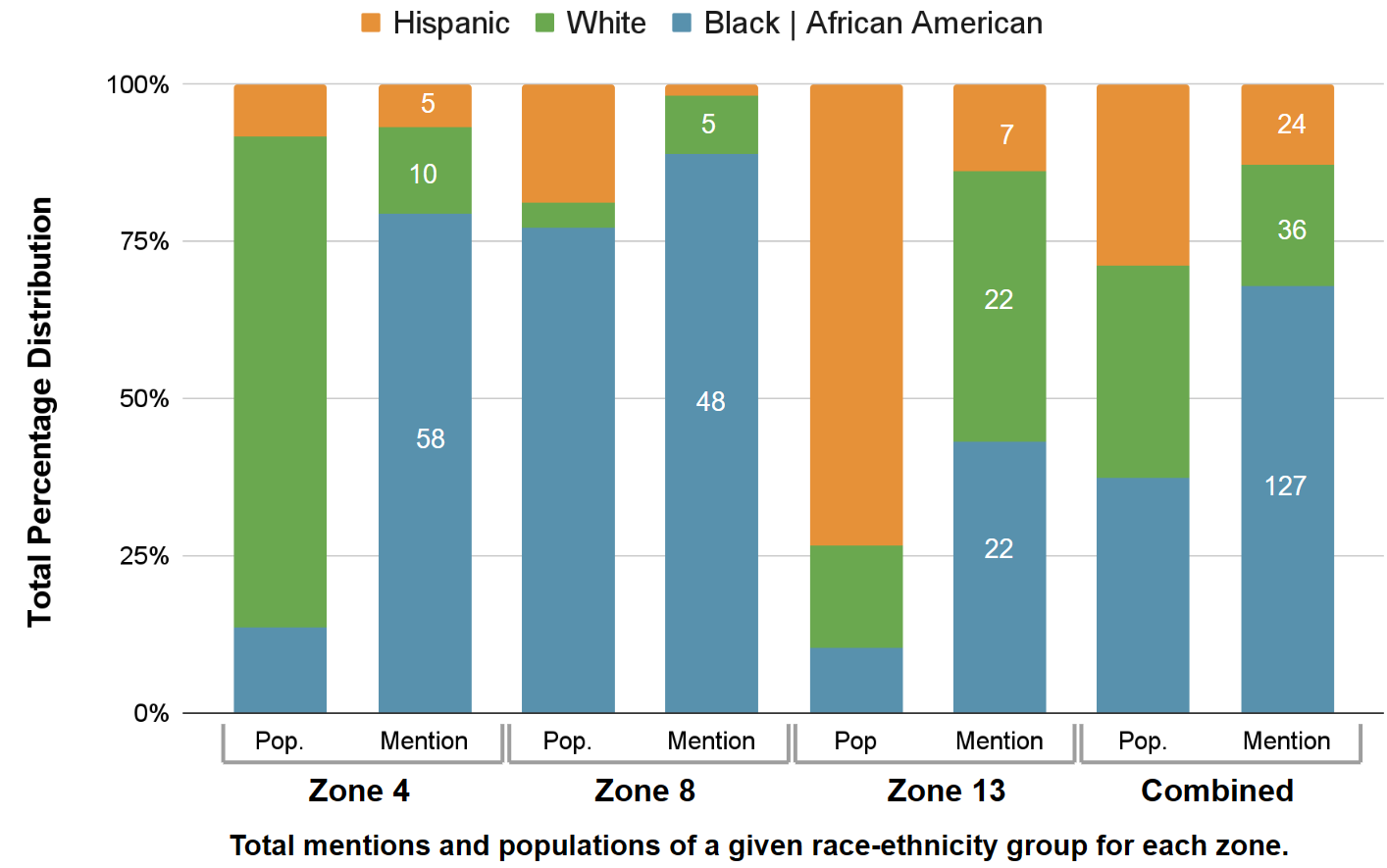}
\caption{Percentage distribution of the total population and the mentions in BPC utterances of various racial/ethnic groups for each zone. The frequency of utterance is mentioned in white.} \label{fig:racementions}
\end{figure}

To investigate communication of developmental status using mentions of age, we ask how many utterances mention minors (less than or equal to 18 years in USA)\footnote{https://www.law.cornell.edu/wex/minor}, which is shown in Fig. \ref{fig:agedist}. Here we see Black individuals have the most mentions of age, including minors. To further explore mentions of age, we categorized individuals into six groups based on the categories developed by the National Institution of Health (NIH) \cite{nih_2022}. These groups included Infants (>1), Children (2-12), Adolescent/Teenager (13-18), Young Adult (19-24), Adult (25-64), and Older Adult (>64). Although age descriptors were only present in 3\% of Event Information Transmissions, we analyzed the distribution of age groups in the utterances that did contain them, as shown in Table \ref{table:age}. Notably, African American or Black \textit{adults} were most frequently mentioned in BPC communication. Furthermore, among utterances that mentioned both age and race, Black and African American ethnicity dominated the Young Adult age group. We also note that among the sociodemographic parameters in consideration, BPC contained relatively few mentions of age. This highlights a potential issue with how officers and dispatchers identify individuals, specifically emphasizing the significant risks that arise from failing to identify a person's developmental status or misidentifying them (e.g., age) when responding to service calls.

Here we present mentions of sociodemographic characteristics as an attention disparity. To the extent minors are a protected class of individuals, this may explain why we observe so few mentions of age. However, to not mention age when a minor is involved places that child at greater risk since responding officers may assume an adult is involved and misinterpret the child's behavior. It is also notable there was not a single mention of White children in BPC across all three dispatch zones (Fig. \ref{fig:agedist}). We are unable to explain this finding but we hypothesize either the age of White children is not conveyed, as we find mentions of age relatively rare in general, or officers rarely communicate about interactions with White children.

\begin{table}[]
\centering
\footnotesize
\begin{tabular}{|c|c|c|}
\hline
\textbf{Race/Ethnicity} & \textbf{Age Range} & \textbf{Utterance Freq.} \\ \hline
 & 2-12 & 3 \\ \cline{2-3} 
Unmentioned & 13-18 & 3 \\ \cline{2-3} 
 & 19-24 & 5 \\ \cline{2-3} 
 & 25-64 & 11 \\ \cline{2-3} 
 & >64 & 3 \\ \hline
& 13-18 & 3 \\ \cline{2-3} 
Black  & 19-24 & 6 \\ \cline{2-3} 
 & 25-64 & 16 \\ \hline
Hispanic & 13-18 & 1 \\ \cline{2-3} 
 & 25-64 & 1 \\ \hline
White & >64 & 9 \\ \hline
\end{tabular}
\caption{Frequency of utterances with mentions of race/ethnicity and age. Age range with no mentions are omitted from the table.} \label{table:age}
\end{table}

\textbf{Privacy vulnerability and sharing of protected health information:} We now explore privacy vulnerability by examining the types of protected health information (PHI) revealed in BPC and the impact of such exposure on each race/ethnicity group. Specifically, we investigate whether there are differences in how each individual directly or indirectly is vulnerable with this system.

Prior research \cite{nissenbaum2009privacy, britz1996technology, solove2004digital} has established the increasing threat to privacy caused by technology, owing to its capability to gather, scrutinize, and circulate personal information. Lyon \cite{lyon2006theorizing, lyon2003surveillance} further emphasizes this notion by defining surveillance, which highlights the \textit{distribution of personal information by technology}. This can lead to technology wielding control over a particular group, and if not managed appropriately, can be misused by external agents \cite{vargas2019digital, schwartz2011pii}. Thus, comprehending the vulnerability of the privacy of a populace is of utmost significance. Drawing from the domains of surveillance, privacy, and digital vulnerability, we analyze the notion of privacy vulnerability. Specifically, from prior research \cite{lyon2006theorizing, vargas2019digital}, we explore \textit{privacy vulnerability} of a demographic group as the probability of their personal identifiable or health-related information being exposed by technology to a third-party. We now examine the existence of privacy vulnerability among various gender, racial/ethnic, and age groups within the three zones of Chicago, through BPC.

\begin{figure}
\centering
\includegraphics[scale = 0.35]{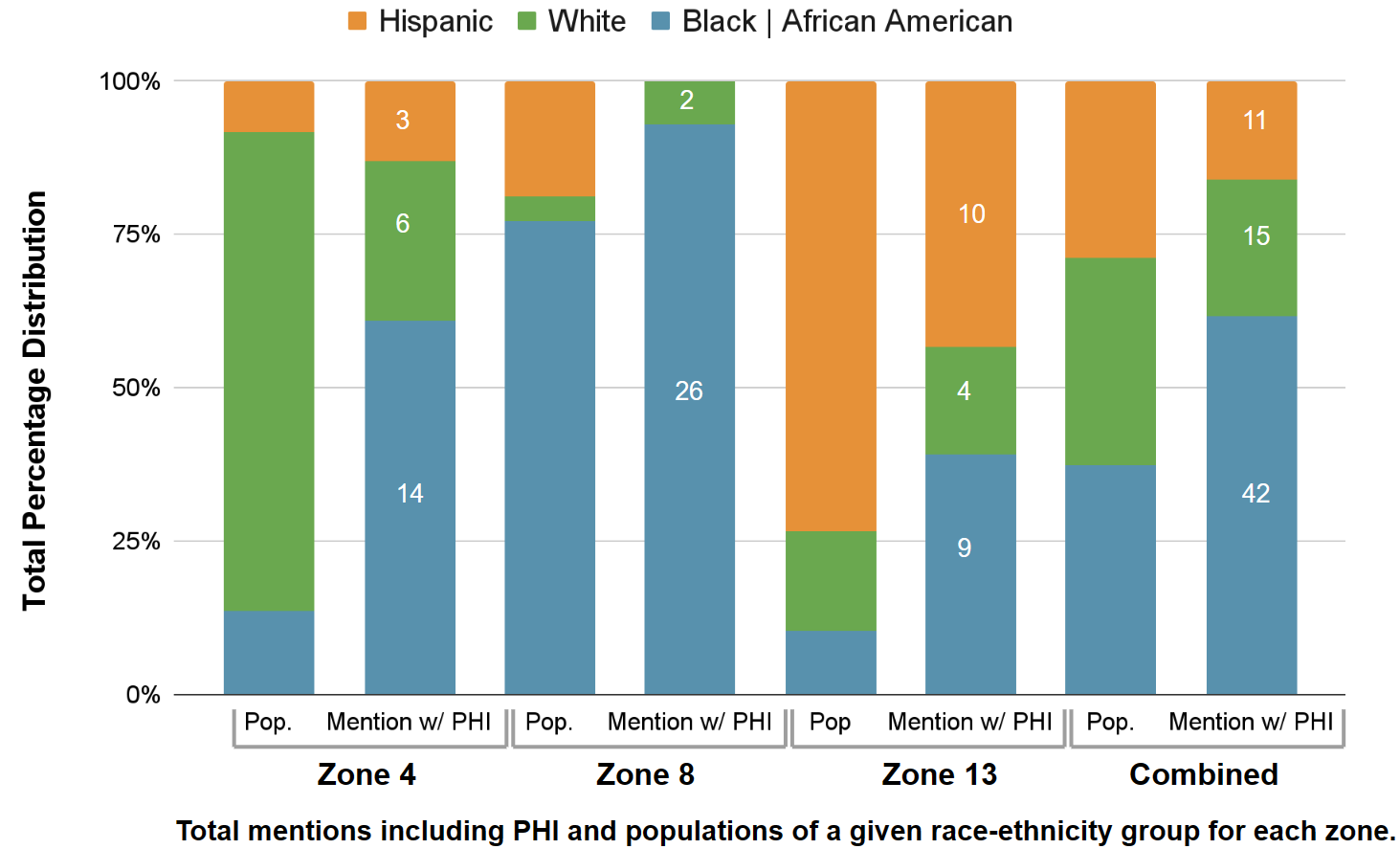}
\caption{Percentage distribution of mentions and PHI in BPC utterances and the total population of various race/ethnicity groups for each zone. The frequency of utterance is mentioned in white.} \label{fig:phimentions}
\end{figure}

To gauge the significance of our findings, we measured privacy vulnerability across all zones for various race/ethnicity groups. 
Fig. \ref{fig:phimentions} denotes the comparison between the percentage of utterance containing PHI and race/ethnicity terms in BPC, and the percent population of that race/ethnicity group in the specific zone. Similar to \textit{technological attention}, our results suggest people of color tend to experience 
greater privacy vulnerability. 

Specifically, we find that the Black and African American populations across all three zones combined are the most vulnerable to privacy misuse and loss (with 61\% of PHI associated to this social group across all three zones).
This observation aligns with the pronounced attention and mentions, in BPC, of the Black population across the selected dispatch zones. On further normalizing the occurrences of PHI mentions to the total number of utterances involving individuals from each racial group, we observe seemingly consistent ratios across all groups (White: 0.45, Hispanic: 0.41, Black: 0.33). This suggests a potential absence of overt behavioral changes when discussing PHI within specific racial communities alone. However, it is also essential to consider that a substantial percentage of these utterances pertain specifically to Black and African American individuals, spotlighting a noteworthy concern regarding the relation between privacy vulnerability and attention disparity. 
This result highlights how disparity in attention can also lead to privacy vulnerability and harm.

We also observe variations in privacy vulnerability for different racial and ethnic groups across the three zones. For instance, in Zone 4 and Zone 13, Black and African American populations are the most vulnerable 
, but in Zone 8, the White population is disproportionately more vulnerable than other groups. 
The common pattern in these two analyses is that the minority population in that zone is most prone to having their personal information disclosed through BPC.

\subsection{Study 4: Test scaling of BPC privacy risk given emerging technologies}

In the current landscape of evolving technological advancements, particularly in LLMs, a critical examination of their implications for personal information security, specifically within accessible conversations or texts, has become essential. The rise of privacy-invasive chatbots and AI as a Service (AIaaS) tools adept at extracting sensitive information from diverse textual sources raises concerns, prompting an upswing in research dedicated to unraveling privacy vulnerabilities associated with ongoing technological breakthroughs. Works like \citet{staab2023beyond} reveal a concerning capability to automatically infer diverse personal attributes from unstructured text with high accuracy. This inference, coupled with widespread LLM adoption, significantly reduces the costs of effort in privacy-infringing inferences, enabling adversarial users to perform such tasks beyond previous constraints imposed by expensive human profilers,

\begin{figure}
\centering
\includegraphics[scale = 0.35]{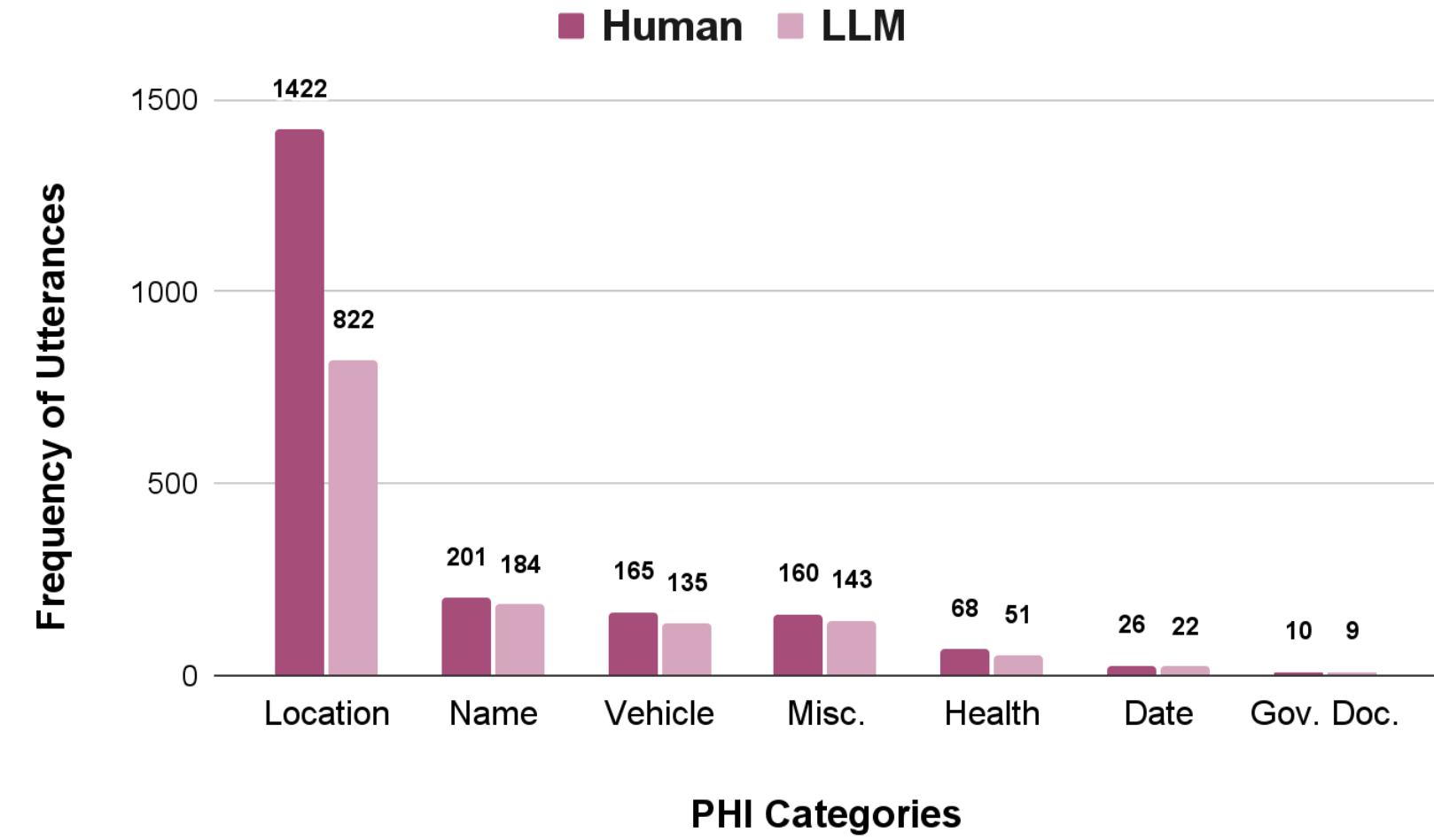}
\caption{Recall of accurately identifying utterances containing PHI by GPT3.5 compared to human annotations.} \label{fig:LLM_Results}
\end{figure}

Therefore, motivated by this concern, we conduct an empirical analysis of BPC using the lens of AIaaS text generation LLMs to understand their capacity to understand policing language while simultaneously identifying PHI sans any prior training. Our choice of the 'off-the-shelf' GPT-3.5 model, devoid of additional finetuning, serves the dual purpose of, first, understanding if we can perform the sensitive task of PHI identification using LLMs and, second, evaluating its proficiency in PHI identification.

To enable the model to perform the task, we use identical instructions used by human annotators to construct the prior gold standard dataset for PHI present in BPC. The prompt is structured as \textit{<Imagine you are a highly intelligent model intended to identify personally identifiable information in Broadcast Police Communications>}, followed by specific details about each PHI group defined for the experiment. This task follows zero-shot prompt formatting to understand the model's ability without additional information. Our deliberate omission of prompt hacking or engineering underscores the model's ability to execute sensitive and privacy-vulnerable tasks with no filter. This raises apprehensions about the model's potential misuse and the inherent risks associated with its capacity to perform such functions with minimal barriers.

The model achieves an accuracy of \textbf{66.5\%} for the overarching PHI identification task from the consolidated set of gold-standard utterances developed by human annotators. This seems relatively diminutive, but further insights into its performance are in Figure \ref{fig:LLM_Results}, showcasing the model's predictions for each PHI group. While the LLM exhibits suboptimal performance in predicting location information, achieving an accuracy of \textbf{86.3\%} for the rest of PHI groups signifies high proficiency in identifying sensitive personal information from BPC without user-end training. Notably, the task's simplicity, requiring no prior skills from users, emphasizes the model's ease of deployment and, therefore, a concerning behavior of demonstrating the ease with which we could extract sensitive information with no additional training. The results also highlight the LLM's notable accuracy in identifying names, health information, and government documents with very high accuracy. Rigorous validation of results is ensured through three separate task executions with similar prompts, yielding consistent outcomes.

These findings unveil the disconcerting potential of these LLMs, enabling adversaries to scale operations far beyond the limitations of expensive human profilers, particularly in extracting sensitive information from ostensibly unstructured data sources 
like BPC as speech technologies improve. The heightened concern arises from the inherent sensitivity of such data sources, which can lead to the inadvertent sharing of personal information given that LLMs can now be used to infer personal information from text corpora or audio corpora without any form of training associated with the same (Eg.: extracting an individuals government based document information or full name from BPC transcripts using publicly available LLM models). 
Moreover, the tendency of BPC to exhibit disparate attention toward minority populations exacerbates their privacy vulnerability to targeted attacks, amplifying the ethical \cite{gautam2024melting} and security implications \cite{staab2023beyond} of both the vulnerable behaviors of BPC and deploying LLMs in such contexts.

\section{Discussion}

Identifying disparities in policing is a complex and multifaceted task that warrants researchers and policymakers' attention. While some empirical investigations \cite{johnson2019officer} have yielded no evidence of anti-Black or anti-Hispanic disparities in police shootings, extant literature \cite{brunson2018kids} has underscored systemic inequality in law enforcement decisions concerning stopping, searching, and arresting minority juveniles. Moreover, works have \cite{chen2021smartphone, geller2017policing} demonstrated that police officers allocate more time to patrolling predominantly Black neighborhoods, engendering exposure disparities that plausibly engender racial disparities in arrest rates. 

Previous research \cite{wells1997research} on the analysis of BPC concluded it is too resource-intensive for research purposes. Given advancements in technology and the application of quantitative analysis in natural language processing, there exists an opportunity to comprehensively investigate various facets of this technology. Notably, \citet{vargas2019digital} conducted a comparative case study on a similar policing technology, exploring the interaction between police officers and dispatchers over public radio frequencies in Chicago. While introducing the concept of digital vulnerability in policing technology, \citet{vargas2019digital} highlighted how certain social groups are more susceptible based on their interactions with these technologies. However, these prior works did not specifically delve into the analysis of BPC nor address the associated privacy vulnerabilities, creating a notable research gap in understanding the broader implications of this technology.

This study, hence, explores the impact of technology utilized by officers and dispatchers, specifically focusing on BPC. It aims to uncover valuable insights into policing behavior while highlighting disparities related to privacy and surveillance. Our findings reveal that (i) we can achieve additional policing insights through the analysis of the technology used, (ii) radio communication in law enforcement agencies, particularly the Chicago Police Department, is susceptible to privacy breaches and can exacerbate social inequalities, and (iii) current technologies like LLMs can easily tap into this technology to extract vulnerable personal information with no filter easily. Analyzing this technology used by dispatchers and officers in the radio communication system can therefore reveal potential disparities against different demographic groups as well as their associated risks.

By employing lexical and thematic qualitative coding, we gain a comprehensive understanding of the intention and policing behavior through communications that occur in BPC. This illuminates the primary inquiry presented in our research. Our descriptive analysis indicates that Black and African American males receive substantially higher communication attention across all four zones, indicating possible racial bias in policing behavior. This result is particularly concerning in Zone 4, a predominantly white area of the city, where communication shows even more attention being trained on Black men. Qualitative coding of text data reveals that most communications in the radio system contain descriptions or mentions of an event and activities, predominantly spoken using police jargon. 

Furthermore, our analysis identifies that 20\% of the utterances in the radio communication system contain PHI, posing a privacy vulnerability. Black and African American communities are more susceptible to privacy vulnerability through radio communication, with approximately 60\% of the utterances containing sensitive information revealing an individual who is Black or African American. These results demonstrate that radio communication in law enforcement agencies can compromise personal privacy and exacerbate existing social inequalities. Furthermore, we extend our analysis to underscore the relevance of this issue in the current era of technological advancement. We demonstrate the ease with which personal information can be extracted from BPC utilizing publicly accessible text generation models, such as GPT3.5.

It is established in the literature that a significant portion of the US population can be uniquely identified through a limited set of attributes, including location, gender, and date of birth \cite{staab2023beyond, sweeney2002k}. This susceptibility poses a potential threat, as malicious actors can exploit this to associate highly personal details inferred from online data, such as mental health status, with real individuals. The implications extend to undesirable or illegal activities, including targeted political campaigns, automated profiling, or stalking. Prior research \cite{vargas2016wounded} has qualitatively highlighted instances wherein policing technologies like BPC attract malicious entities engaged in eavesdropping activities to discern policing movements and surveil individuals of interest. The works of \citet{staples1997culture} and \citet{lyon2003surveillance} have also laid the groundwork for the concept of the \textit{watchful gaze} of technology and its implications for surveillance in social settings. Hence, it is crucial to comprehend the existence of personal information within technologies like BPC to understand their societal ramifications. Our findings, hence, extend the awareness on the potential disparities embedded in such technological interventions and also emphasize the privacy concerns associated with them.

The privacy vulnerabilities in these technologies are further exacerbated with the advent of new technologies, such as LLMs, which are publicly accessible and capable of executing malicious tasks without adequate filters. Prior works have already demonstrated how societal bias is now an automated behaviour \cite{venkit2023automated, gupta2023survey, venkit2022study} through these models and why we need to be cautious of their ramifications \cite{dev2021measures}. With such technology, we now can enable automation of privacy vulnerability. Our results substantiate this concern. Despite these models achieving performance levels approaching expert human capabilities, they do so at a significantly reduced cost, necessitating 100 times less financial and 240 times lower time investment than human labelers \cite{staab2023beyond}. This cost-effectiveness renders privacy violations on a large scale feasible for the first time. Thus, this work underscores the imperative of scrutinizing the language employed in law enforcement communication technologies such as BPC. Our findings therefore provide, first-of-its-kind, insights for future research concerning the language dynamics within law enforcement communication systems and their implications for policing behavior and social inequality in regards to the current technological environment.

\section{Conclusion}

In this study, we explored the interaction between policing technology and society, focusing on the influence of BPC, on policing behavior and related social inequalities in Chicago. Our contribution advances research on this topic by unpacking the societal impact of BPC via its implications for privacy vulnerability, a dimension largely unexplored in prior research, and disparities in police attention. We found 
disproportionate attention given to Black and African American males in communication patterns, even within predominantly White geographic areas. Additionally, we identified critical privacy vulnerabilities within BPC, wherein a substantial portion of radio communications contained sensitive information, including personal health information.

Lexical analysis reveals that BPC transmissions are typically brief in word count, and they often contain abbreviated codes for common words and phrases. Our descriptive analysis shows substantially more frequent communication regarding Black males compared to other demographic groups across all three zones, with Zone 4, a predominantly White area of the city, showing that more attention is given to Black males than all other permutations of gender and race/ethnicity combined. Our qualitative coding of BPC identifies specific categories that differentiate the communications based on their ostensible purpose and function. We also identify that one out of every five utterances in BPC reveal some form of Personal Health Information.

Our analysis reveals a concerning trend of racial inequality in terms of the exposure of sensitive information during conversations. Specifically, our findings suggest that Black communities are at a higher risk of privacy vulnerability through BPC. We see that $\sim$60\% of the utterances (that contain both race/ethnicity and PHI) contain sensitive information revealed of an individual who is Black or African American (particularly in zones where they constitute the minority), showing strong suggestions of disparate privacy vulnerability in BPC. The importance of this analysis gains added significance when employing LLMs to explore the feasibility of capturing information effortlessly. A new form of technological interaction emerges as millions of people have started interacting with thousands of custom chatbots and text generation models \cite{staab2023beyond}. Understanding how LLMs can enhance potential vulnerability in technologies like BPC is required. Our findings underscore the concerning impact, as LLMs, without any additional training, demonstrated an ability to capture PHI with an accuracy exceeding 85\%. This high level of accuracy raises alarm regarding the vulnerability of PHI within BPC.

Significantly, our findings underscore the susceptibility of Black and African American communities to privacy breaches. The gravity of these results is underscored by the use of current technology, specifically large language models, to identify the presence of sensitive data in BPC transcripts. This study thus highlights the need to address vulnerabilities in underexplored technologies, like BPC, as such
characteristics of this technology can result in misuse by nefarious actors. 

As a novel exploration of policing behavior through the lens of BPC and privacy, our study contributes to the existing literature on policing technology \cite{brayne_big_2017, braga_does_2022, goldberg_how_2022, chapman_data-driven_2022, ratcliffe_philadelphia_2021}. The revelations from this research not only lay the groundwork for subsequent inquiries into language usage within law enforcement communication systems but also offer broader insights into the societal ramifications of such systems. This study, therefore, aids in understanding the utilization of technology in law enforcement, emphasizing the indispensable necessity for ethical and responsible practices.

\section{Limitations and Future Work}

Our study, framed within the context of CSCW, has limitations. First, analysis is constrained by its exclusive focus on the Chicago Police Department as a case study. This limitation restricts our ability to generalize study findings, as policing practices and communication dynamics may vary across police systems with different operational structures and officer/resident demographics.

Second, our research is limited to three prominent dispatch zones within Chicago, capturing communication patterns within a confined spatial and temporal scope. Due to the taxing nature of data collection, we were only able to study the time frame used here. Other factors such as local policies, historical context, and community dynamics, which influence communication practices, could not be incorporated into our current analysis, though we intend to do so in future work.
These constraints highlight avenues for future research to study police departments across the country and opportunities to incorporate a broader range of contextual factors, ensuring a more comprehensive understanding of communication dynamics in policing.

This study is an essential milestone in an ongoing project of understanding and analyzing BPC. In the upcoming work, we intend to demonstrate the extent to which fine-tuning improves the performance of speech recognition models, train neural-based models to perform speaker diarization, and detect conversations about the same incident interspersed with conversations about other incidents. These efforts must be understood as further explorations of risk to identify best practices and ethical standards for model deployment. Creating these sociotechnical AI safety artifacts will ensure BPC's potential risks can be addressed before widespread research use of BPC using tools under development \cite{lazar_ai_2023}.

\begin{acks}
Research supported by National Institute of Minority Health and Health Disparities of the National Institutes of Health under award number R01MD015064. Content does not necessarily represent the official views of the National Institutes of Health. Research also supported by the Urban Resiliency Initiative. The authors would like to thank Margaret Beale Spencer for her extensive support and critical feedback. 
\end{acks}

\bibliographystyle{ACM-Reference-Format}

\bibliography{sample-base}


\end{document}